\documentclass[10pt,twocolumn,letterpaper]{article}

\usepackage[pagenumbers]{cvpr} % To force page numbers, e.g. for an arXiv version

\usepackage{graphicx}
\usepackage{amsmath}
\usepackage{amssymb}
\usepackage{booktabs}
\usepackage{multirow}
\usepackage{url}
\usepackage[pagebackref,breaklinks,colorlinks]{hyperref}
\usepackage{appendix}
\usepackage{setspace}

\usepackage[capitalize]{cleveref}
\crefname{section}{Sec.}{Secs.}
\Crefname{section}{Section}{Sections}
\Crefname{table}{Table}{Tables}
\crefname{table}{Tab.}{Tabs.}

\def\confName{CVPR}
\def\confYear{2023}

\begin{document}

\title{Super-Resolution Neural Operator}

\author{
Min Wei\thanks{Equal contributions. This work is supported by the National Natural Science Foundation of China (61871055).}\qquad Xuesong Zhang$^\ast$\thanks{Corresponding author.}\\
Beijing University of Posts and Telecommunications, China\\
{\tt\small \{mw, xuesong\underline{~}zhang\}@bupt.edu.cn}
}

\maketitle

%%%%%%%%% ABSTRACT
\begin{abstract}
   We propose Super-resolution Neural Operator (SRNO), a deep operator learning framework that can resolve high-resolution (HR) images at arbitrary scales from the low-resolution (LR) counterparts. Treating the LR-HR image pairs as continuous functions approximated with different grid sizes, SRNO learns the mapping between the corresponding function spaces. From the perspective of approximation theory, SRNO first embeds the LR input into a higher-dimensional latent representation space, trying to capture sufficient basis functions, and then iteratively approximates the implicit image function with a kernel integral mechanism, followed by a final dimensionality reduction step to generate the RGB representation at the target coordinates. The key characteristics distinguishing SRNO from prior continuous SR works are: 1) the kernel integral in each layer is efficiently implemented via the Galerkin-type attention, which possesses non-local properties in the spatial domain and therefore benefits the grid-free continuum; and 2) the multilayer attention architecture allows for the dynamic latent basis update, which is crucial for SR problems to ``hallucinate" high-frequency information from the LR image. Experiments show that SRNO outperforms existing continuous SR methods in terms of both accuracy and running time. Our code is at \small\url{https://github.com/2y7c3/Super-Resolution-Neural-Operator}
\end{abstract}

%%%%%%%%% BODY TEXT
\section{Introduction}
\label{sec:intro}

\begin{figure}
\centering
\includegraphics[scale = 0.4]{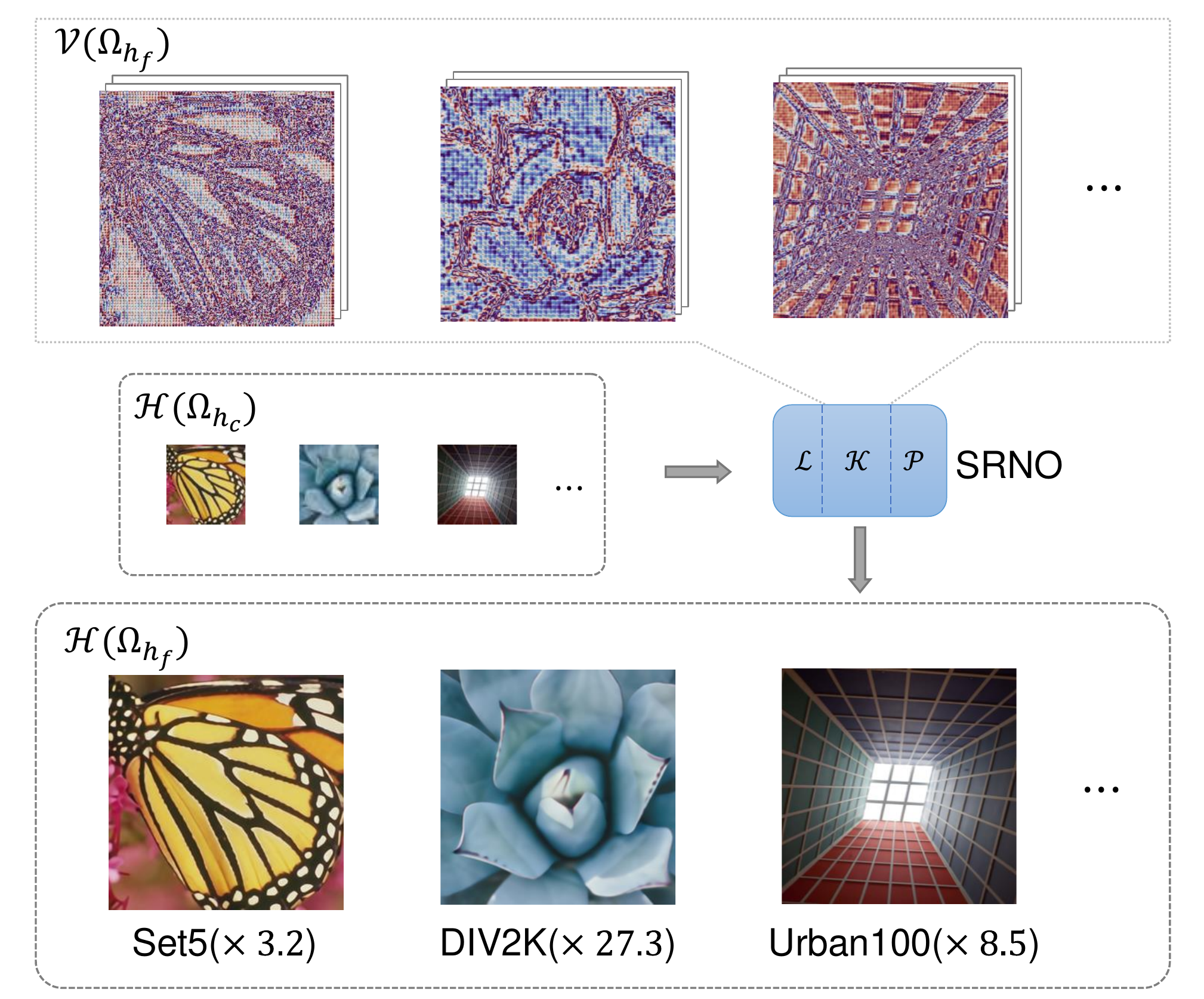}
\caption{\textbf{Overview of Super-Resolution Neural operator (SRNO)}. SRNO is composed of three parts, $\mathcal{L}$ (Lifting), $\mathcal{K}$ (kernel integrals) and $\mathcal{P}$ (Projection), which perform consecutively to learn mappings between approximation spaces $\mathcal{H}(\Omega_{h_c})$ and $\mathcal{H}(\Omega_{h_f})$ associated with grid sizes $h_c$ and $h_f$, respectively. The key component, $\mathcal{K}$, uses test functions in the latent Hilbert space $\mathcal{V}(\Omega_{h_f})$ to seek instance-specific basis functions.}
\label{fig:srnoover}
\end{figure}

Single image super-resolution (SR) addresses the inverse problem of reconstructing high-resolution (HR) images from their low-resolution (LR) counterparts. In a data-driven way, deep neural networks (DNNs) learn the inversion map from many LR-HR sample pairs and have demonstrated appealing performances~\cite{chen2021pre,9607618,lim2017enhanced,zhang2018image,zhang2018residual,shi2016real,magid2021dynamic}. Nevertheless, most DNNs are developed in the configuration of single scaling factors, which cannot be used in scenarios requiring arbitrary SR factors~\cite{wang2021learning,yang2021implicit}. Recently, implicit neural functions (INF)~\cite{chen2021learning,lee2022local} have been proposed to represent images in arbitrary resolution, and paving a feasible way for continuous SR. These networks, as opposed to storing discrete signals in grid-based formats, represent signals with evaluations of continuous functions at specified coordinates, where the functions are generally parameterized by a multi-layer perceptron (MLP). To share knowledge across instances instead of fitting individual functions for each signal, encoder-based methods~\cite{chen2021learning,park2019deepsdf,jiang2020meshfreeflownet} are proposed to retrieve latent codes for each signal, and then a decoding MLP is shared by all the instances to generate the required output, where both the coordinates and the corresponding latent codes are taken as input. However, the point-wise behavior of MLP in the spatial dimensions results in limited performance when decoding various objects, particularly for high-frequency components~\cite{tancik2020fourier,sitzmann2020implicit}.  

Neural operator is a newly proposed neural network architecture in the field of computational physics~\cite{lu2021learning,li2020fourier,kovachki2021neural} for numerically efficient solvers of partial differential equations (PDE). Stemming from the operator theory, nueral operators learn mappings between infinite-dimensional function spaces, which is inherently capable of continuous function evaluations and has shown promising potentials in various applications~\cite{pathak2022fourcastnet,guibas2021efficient,hwang2022solving}. Typically, neural operator consists of three components: 1) lifting, 2) iterative kernel integral, and 3) projection. The kernel integrals operate in the spatial domain, and thus can explicitly capture the global relationship constraining the underlying solution function of the PDE. The attention mechanism in transformers~\cite{vaswani2017attention} is a special case of kernel integral where linear transforms are first exerted to the feature maps prior to the inner product operations~\cite{kovachki2021neural}. Tremendous successes of transformers in various tasks~\cite{dosovitskiy2020image, liu2021swin, touvron2021training} have shown the importance of capturing global correlations, and this is also true for SR to improve performance.~\cite{gu2021interpreting}.

In this paper, we propose the super-resolution neural operator (SRNO), a deep operator learning framework that can resolve HR images from their LR counterparts at arbitrary scales. As shown in Fig.\ref{fig:srnoover}, SRNO learns the mapping between the corresponding function spaces by treating the LR-HR image pairs as continuous functions approximated with different grid sizes. The key characteristics distinguishing SRNO from prior continuous SR works are: 1) the kernel integral in each layer is efficiently implemented via the Galerkin-type attention, which possesses non-local properties in the spatial dimensions and have been proved to be comparable to a Petrov-Galerkin projection~\cite{cao2021choose}; and 2) the multilayer attention architecture allows for the dynamic latent basis update, which is crucial for SR problems to “hallucinate” high-frequency information from the LR image. When employing same encoders to capture features, our method outperforms previous continuous SR methods in terms of both reconstruction accuracy and running time.

In summary, our main contributions are as follows:
\begin{itemize}
    \item We propose the methodology of super-resolution neural operator that maps between finite-dimensional function spaces, allowing for continuous and zero-shot super-resolution irrespective the discretization used on the input and output spaces.
    \item We develop an architecture for SRNO that first explores the common latent basis for the whole training set and subsequently refines an instance-specific basis by the Galerkin-type attention mechanism. 
    \item Numerically, we show that the proposed SRNO outperforms existing continuous SR methods with less running time, and even generates better results on the resolutions for which the fixed scale SR networks were trained.
\end{itemize}

%----------------------------------------------------------
\section{Related Work}

\textbf{Deep learning based SR methods}.
\cite{chen2021pre,9607618,lim2017enhanced,zhang2018image,zhang2018residual,shi2016real,magid2021dynamic} have achieved impressive performances, in multi-scale scenarios one has to train and store several models for each scale factor, which is unfeasible when considering time and memory budgets. In recent years, several methods~\cite{hu2019meta,wang2021learning,son2021srwarp} are proposed to achieve arbitrary-scale SR with a single model, but their performances are limited when dealing with out-of-distribution scaling factors. Inspired by INF, LIIF~\cite{chen2021learning} takes continuous coordinates and latent variables as inputs, and employs an MLP to achieve outstanding performances for both in-distribution and out-of-distribution factors. In contrast, LTE~\cite{lee2022local} transforms input coordinates into the Fourier domain and uses the dominant frequencies extracted from latent variables to address the spectral bias problem~\cite{tancik2020fourier,sitzmann2020implicit}. In a nutshell, treating images as RGB-valued functions and sharing the implicit function space are the keys to the success of LIIF-like works~\cite{chen2021learning,lee2022local}. Nevertheless, a purely local decoder, like MLP, is not able to accurately approximate arbitrary images, although it is rather sensitive to the input coordinates.

\begin{figure*}
\centering
\includegraphics[scale = 0.5]{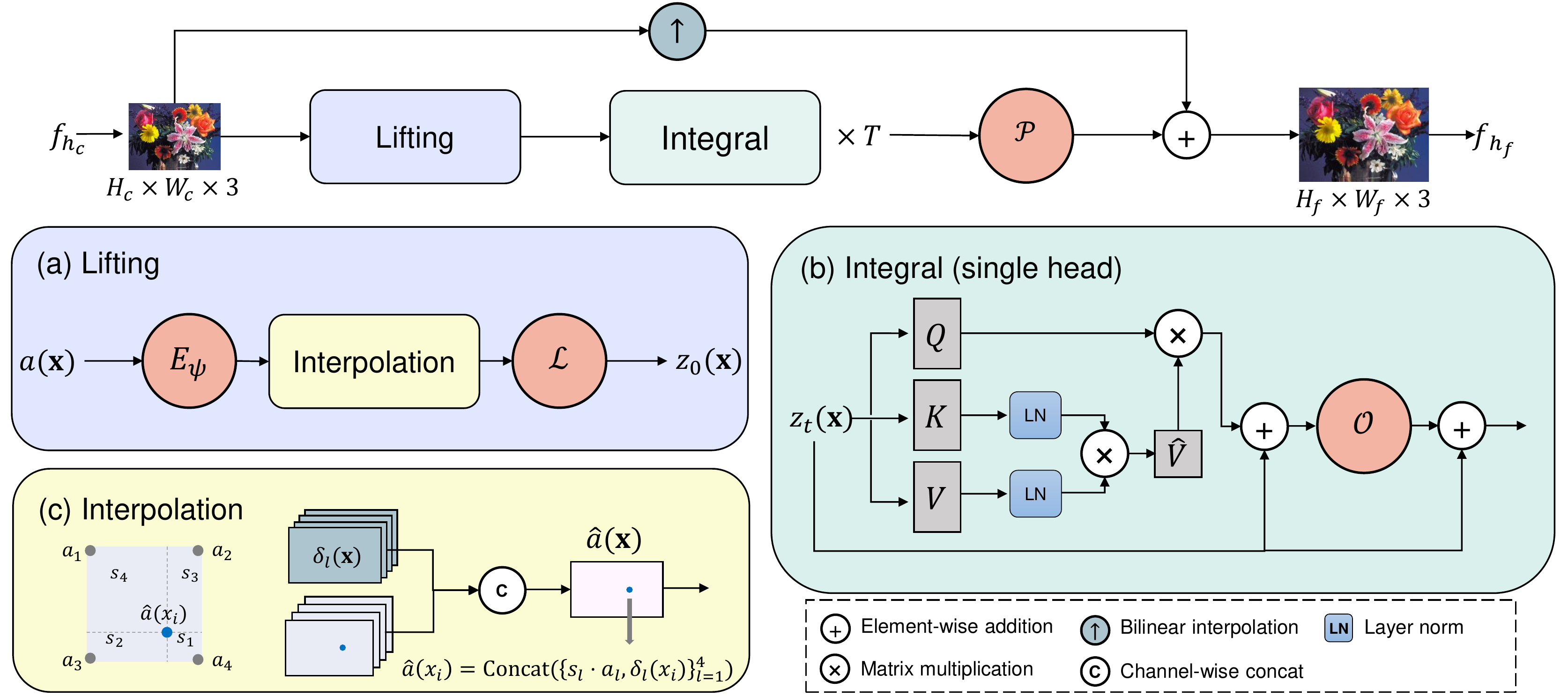}
\caption{\textbf{Super-resolution neural operator (SRNO) architecture for continuous SR.} The input LR image $f_{h_c}$ undergoes three phases to output the HR image $f_{h_f}$ with the specified resolution: (a) Lifting the LR pixel values $a(\textbf{x})$ on the set of coordinates $\textbf{x}=\{x_i\}_{i=1}^{n_{h_f}}$ to a higher dimensional feature space by a CNN-based encoder $E_\psi$, constructing the latent representation $\hat{a}(x)$, and linearly transforming into the first layer's input $z_0(\textbf{x})$. (b) kernel integrals composed of $T$ layers of Galerkin-type attention, and (c) finally project to the RGB space.}
\label{fig:srno}
\end{figure*}

\textbf{Neural Operators}. Recently, a novel neural network architecture, Neural Operator (NO), was proposed for discretization invariant solutions of PDEs via infinite-dimensional operator learning~\cite{li2020fourier,gupta2021multiwavelet,lu2021learning,kovachki2021neural}. Neural operators only need to be trained once and are capable of transferring solutions between differently discretized meshes while keeping a fixed approximation error. A valuable merit of NO is that it does not require knowledge of the underlying PDE, which allows us to introduce it by the following abstract form,
\begin{equation}
\label{no}
\begin{split}
    (\mathbf{L}_au)(x) &= f(x), \quad x\in D,\\
    u(x)&=0, \quad x\in \partial D,
\end{split}
\end{equation}
where $u : D\rightarrow \mathbb{R}^{d_u}$ is the solution function residing in the Banach space $\mathcal{U}$, and $\mathbf{L} : \mathcal{A} \rightarrow \mathbf{L} (\mathcal{U}; \mathcal{U}^*)$ is an operator-valued functional that maps the coefficient function $a\in \mathcal{A}$ of the PDE to $f\in \mathcal{U}^{*}$, the dual space of $\mathcal{U}$. As in many cases the inverse operator of $\mathbf{L}$ even does not exist, NO seeks a feasible operator $\mathcal{G}: \mathcal{A}\rightarrow\mathcal{U}, a \mapsto u$, directly mapping the coefficient to the solution within an acceptable tolerance. 

The operator $\mathcal{G}$ is numerically approximated by training a neural network $\mathcal{G}_{\theta} : \mathcal{A}\rightarrow\mathcal{U}$, where $\theta$ are the trainable parameters. Suppose we have $N$ pairs of observations $\{a_j,u_j\}_{j=1}^N $ where the input functions $a_j$ are sampled from probability measure $\mu$ compactly supported on $\mathcal{A}$, and $u_j = \mathcal{G}(a_j)$ are used as the supervisory output functions. The infinitely dimensional operator learning problem $\mathcal{G}\leftarrow \mathcal{G}_\theta$ thus is associated with the empirical-risk minimization problem~\cite{vapnik1998statistical}. In practice, we actually measure the approximation loss using the sampled observations $u_o^{(j)}$ and $a_o^{(j)}$, which are the direct results of discretization:
\begin{equation}
\begin{split}
 &\min_{\theta}{\mathbb{E}_{a\sim\mu}\lVert\mathcal{G}(a)-\mathcal{G}_\theta(a)\rVert_\mathcal{U}} \\
 &\approx \min_{\theta}\frac{1}{N}\sum_{j=1}^N{\lVert u_o^{(j)} - \mathcal{G}_{\theta}(a_o^{(j)}\rVert}_{\mathcal{U}}.
\end{split}
\end{equation}

Similar to classical feedforward neural networks (FFNs), the NO is of an iterative architecture. For ease of exposition, suppose $\mathcal{A}$ is defined on the bounded domain $D\subset \mathbb{R}^d$, and the inputs and outputs of the intermediate layers are all vector-valued functions, with dimension $d_z$. Then $\mathcal{G}_{\theta}: \mathcal{A} \rightarrow \mathcal{U}$ can be formulated as follows:
\begin{align}
 z_0(x) &= \mathcal{L}(x,a(x)), \label{lifing}  \\
 z_{t+1}(x) &= \sigma(W_{t}z_{t}(x) + (\mathcal{K}_t(z_t;\Phi))(x)) ,\label{iterative} \\
 u(x) &= \mathcal{P}(z_T(x)),
\end{align}
where $\mathcal{L}: \mathbb{R}^{d_a+d} \rightarrow \mathbb{R}^{d_z}$, and $\mathcal{P}: \mathbb{R}^{d_z} \rightarrow \mathbb{R}^{d_u}$ are the local lifting and projection functions respectively , mapping the input $a$ to its first layer hidden representation $z_0$ and the last layer hidden representation $z_T$ back to the output function $u$. $W: \mathbb{R}^{d_z} \rightarrow \mathbb{R}^{d_z}$ is a point-wise linear transformation, and $ \sigma: \mathbb{R}^{d_z} \rightarrow \mathbb{R}^{d_z}$ is the nonlinear activation function.

Although the PDE in \eqref{no} point-wisely defines the behavior of the solution function $u(x)$, the solution operator $\mathcal{G}$ we are seeking should exhibit the non-local property such that $\mathcal{G}(a)$ can approximate $u$ everywhere rather than locally. For this purpose, NO may employ the kernel integral operators $ \mathcal{K}_t: \{z_t: {D}_t\rightarrow\mathbb{R}^{d_z}\} \mapsto \{{z_{t+1}: {D_{t+1}} \rightarrow \mathbb{R}^{d_z}}\}$ to maintain the continuum in the spatial domain, and one of the most adopted forms~\cite{li2020fourier} is defined as
\begin{equation}
 (\mathcal{K}_t(z_t;\Phi))(x) = \int_{D}{K_t(x,y;\Phi)z_t(y)\mathrm{d}y}, \quad \forall x\in D, \label{kernel}
\end{equation}
where the kernel matrix $K_t: \mathbb{R}^{d+d} \rightarrow \mathbb{R}^{d_z \times d_z}$ is parameterized by $\Phi$.

%-------------------------------------------------------------------------
\section{Super-resolution Neural Operator}

All operations of NO are defined in function space and the training data are just samples at coordinates ${x_i}$ irrespective of the discretization sizes. This property inspires us to employ the NO's archetecture for continuous SR. Different from NO that aims at mapping between infinite-dimensional function spaces, we define SRNO in the sense of mappings between two approximation spaces of finite-dimensional but continuous functions. This is reasonable because imaging optics inevitably limit the highest possible frequency in the radiance field under observation. With this configuration, we can design the archetecture of SRNO using the well-developed theoretical tool, the Galerkin-type method that is widely used in the filed of Finite elements~\cite{ern2004theory}.

\textbf{Problem Formulation}. Let $(\mathcal{H},\langle\cdot,\cdot\rangle_{\mathcal{H}})$ denote a Hilbert space equipped with the inner-product structure, which is continuously embedded in the space of continuous functions $C^0(\Omega)$, with $\Omega \subset \mathbb{R}^2$ a bounded domain. An image is defined as a vector-valued function $\mathcal{H}\ni f:\Omega \rightarrow \mathbb{R}^3$. Assume we can access the function values of $f$ at the coordinates $\{x_i\}_{i=1}^{n_h} \subset \Omega$ with the biggest discretization size $h$. Note that $x_i$'s are not necessarily equidistant. Our goal is to learn a super-resolution neural operator between two Hilbert spaces with different resolutions: $\mathcal{S}_{\theta}: \mathcal{H}\supset\mathcal{H}(\Omega_{h_c}) \rightarrow \mathcal{H}(\Omega_{h_f})\subset\mathcal{H}$, where $h_c,h_f$ denotes the coarse and the fine grid sizes, respectively. Given $N$ function pairs $\{a^{(k)},u^{(k)}\}_{k=1}^N$, where $a^{(k)}\in\mathcal{H}(\Omega_{h_c})$ and $u^{(k)}\in\mathcal{H}(\Omega_{h_f})$. Our SRNO parameterized by $\theta$ can be solved through the associated empirical-risk minimization problem:
\begin{equation}
\setlength{\abovedisplayskip}{3pt}
\setlength{\belowdisplayskip}{3pt}
 \min_{\theta}\frac{1}{N}\sum_{k=1}^N{\lVert u_{h_f}^{(k)}-\mathcal{S}_{\theta}(a_{h_c}^{(k)}) \rVert_{\mathcal{H}}.}
\end{equation}
Since our data $a^{(k)}$ and $u^{(k)}$ are functions, to work with them numerically, we assume access only to their point-wise evaluations. Let $\mathbb{A}_{h_c}\subset\mathcal{H}(\Omega_{h_c}) \subset \mathcal{H}$ be an approximation space~\cite{cao2021choose} associated with $\{x_i\}^{n_{h_c}}_{i=1}$, such that for any $a\in \mathbb{A}_{h_c}$, $a(\cdot) = \sum_{i=1}^{n_{h_c}}{a(x_i)\xi_{x_i}(\cdot)}$, where $\{\xi_{x_i}(\cdot)\}$ form a set of nodal basis for $\mathbb{A}_{h_c}$ in the sense that $\xi_{x_i}(x_j) = \delta_{ij}$. Similarly, $\mathbb{U}_{h_f}\subset\mathcal{H}(\Omega_{h_f}) \subset \mathcal{H}$ is another approximation space associate with grid size $h_f$. Note that Our observed LR-HR image pairs $\{a,u\}$ may have different grid size pairs $\{h_c,h_f\}$, which means they belongs to different approximation spaces $\{\mathbb{A}_{h_c},\mathbb{U}_{h_f}\}$.

\textbf{Lifting}. The neural operators from PDEs usually use a simple pointwise function, $\mathcal{L}$ in (\ref{lifing}), to expand the input channels. But for the SR problem, a deep feature encoder $E_\psi$ (with trainable parameters $\psi$) and spatial interpolations should be considered in the lifting operation, due to the complexity of natural images and the discretization inconsistency between LR and HR image functions. Another critical ingredient in Lifting is the incorporation scheme of coordinates. Prior study~\cite{islam2019much} demonstrates that deep CNNs can implicitly learn to encode the information about absolute positions. As shown in Fig.\ref{fig:srno} (c), the position information of the grid points $\{a_l\}_{l=1}^4$ have been implicitly encoded by $E_\psi$ employing CNNs. Therefore, we only need to explicitly construct the coordinate features $\hat{a}(x)$ with the fractional part of coordinate $x$ inside a grid. In order to reduce the blocky artifacts resulting from direct interpolation of the LR feature maps, we propose to concatenate the features weighted by the corresponding bilinear interpolation factors $s_l$. Compared to the local ensemble trick in \cite{chen2021learning}, our method takes shorter running time and overcomes the over-smoothing problem. We reformulate the lifting operation in (\ref{lifing}) as: 
\begin{equation}
 z_0(x) = \mathcal{L}(\mathbf{c}, \{s_l\cdot E_{\psi}(a(\hat{x}_l)), {\delta}_l(x)\}_{l=1}^4),
\end{equation}
where $x$ are coordinates of HR image functions, $\hat{x}_l$ are the coordinates of the four neighbors of $x$, ${\delta}_l(x)=x-\hat{x}_l$, $\mathbf{c}=(2/r_x,2/r_y)$ represents a $r_x\times r_y$ local area in HR images with $r_x$ and $r_y$ the scaling factors, and $\mathcal{L}: \mathbb{R}^{4\times (d_e+2)+2} \rightarrow \mathbb{R}^{d_z}$ is an local linear transformation function, with $d_e$ the number of channels after $E_\psi$.

\textbf{Kernel integral}. The kernel integral operator, in \eqref{iterative}, actually tells that we can identify the hidden representation $z(x)$ of the underlying image function $f(x)$ with distributions~\cite{rudin1991functional}, inner products in the settings of SRNO. We first define the kernel $K_t:\mathbb{R}^{d_z+d_z} \rightarrow \mathbb{R}^{d_z\times d_z}$ with respect to the input pair $(z(x),z(y))$, rather than like \eqref{kernel} depending on the spatial variables $(x,y)$. Furthermore, in order to more efficiently explore $z(x)$ we can use multiple sets of test functions defining the distributions, which may immediately remind one of a single-head self-attention~\cite{kovachki2021neural}. Denote $z_i=z(x_i)\in \mathbb{R}^{d_z}$ for $i=1,\dots,n_{h_f}$. The kernel integral operator can be approximated by the Monte-Carlo method (omitting the layer index): 
\begin{equation}
\setlength{\abovedisplayskip}{3pt}
\setlength{\belowdisplayskip}{3pt}
\begin{split}
 & (\mathcal{K}(z))(x) = \int_{\Omega}{K(z(x),z(y))z(y)\mathrm{d}y} \\ 
 &\approx \sum_{i=1}^{n_{h_f}}{K(z(x),z_i)z_i}, \quad \forall x\in \Omega_{h_f}, \label{attn}
\end{split}
\end{equation}
where
\begin{equation}
\setlength{\abovedisplayskip}{3pt}
\setlength{\belowdisplayskip}{3pt}
 K(z(x),z_i) = \frac{\exp{(\frac{\langle W_qz(x),W_kz_i\rangle}{\sqrt{d_z}})}}{\sum_{j=1}^{n_{h_f}}{\exp{(\frac{\langle W_qz_j,W_kz_i\rangle}{\sqrt{d_z}})}}}W_{v}.
\end{equation}
In the language of transformers, the matrices $W_q,W_k,W_v\in \mathbb{R}^{d_z\times d_z}$ correspond to the queries, keys and values functions respectively. By allowing $K$ to be the sum of multiple functions with separate trainable parameters, the multi-head self-attention can also be rewritten like in \eqref{attn}. For every observation $(a,u)$, this operation has a complexity of $O(n_{h_f}^2d_z)$, which is unaffordable for the SR problem where the sampling number $n_{h_f}$ usually reaches to $10,000$ or beyond. To overcome this issue, we employ the Galerkin-type attention operator which has a linear complexity $O(n_{h_f}d_z^2)$. The approximation capacity of a linearized Galerkin-type attention has been proved to be comparable to a Petrov-Galerkin projection~\cite{cao2021choose}. 

We suppose that a image function $f$ is locally integrable, i.e., $f$ is measurable and $\int_\Omega{|f|}<\infty$ for every compact $\Omega\subset\mathbb{R}^2$. The idea behind distributions~\cite{rudin1991functional} is to identify $f$ with $\int{f\phi}$ by suitablely choosing a ``test function" $\phi$, rather than by a series of separate evaluations $f(x)$ at coordinates $x\in \Omega$. Let $v(x)=W_vz(x), k(x)=W_kz(x)$, and $q(x)=W_qz(x)$ be the keys, queries and values functions respectively, which are $d_z$ dimensional vector-valued functions, with a subscript denoting the corresponding component. The kernel integral attention then can be written in a component-wise form ($j=1, \dots ,d_z$):
\begin{equation}
\begin{split}
 &\left((\mathcal{K}(z))(x)\right)_j = \sum_{l=1}^{d_z}{\langle k_l, v_j\rangle q_l(x)} \\
 &\approx \sum_{l=1}^{d_z}{\left(\int_{\Omega}{k_l(y)v_j(y)\mathrm{d}y}\right)q_l(x)},\quad \forall x\in \Omega_{h_{f}}. 
\end{split}
\end{equation}

Denote the evaluations $(z(x_1),\dots,z(x_{n_{h_f}}))^T$ by $\mathbf{z}\in \mathbb{R}^{n_{h_f}\times d_z}$. Let $Q=\mathbf{z} W_q , K=\mathbf{z} W_k, V=\mathbf{z} W_v \in \mathbb{R}^{n_{h_f}\times d_z}$. The columns of $Q/K/V$ contain the vector representations of the learned basis functions, spanning certain subspaces of the latent representation Hilbert spaces respectively. 
\begin{equation}
\label{eq:galerkin}
 \mathbf{z} = Q(\widetilde{K}^T\widetilde{V})/n_{h_f}, 
\end{equation}
where $\widetilde{K} = \text{Ln}(K),\widetilde{V} = \text{Ln}(K)$, with $\text{Ln}(\cdot)$ the layer normalization. The $j$-th column of $\widetilde{K}^T\widetilde{V}$ contains the coefficients for the linear combination of the basis vectors $\{q_j\}_{j=1}^{d_z}$ to form the output $\mathbf{z}$. Consequently, the global correlations are reflected through the components of $\widetilde{K}^T\widetilde{V}$. As a result, SRNO can utilize all information effectively without the feature unfolding~\cite{chen2021learning} or local convolutions~\cite{lee2022local}, significantly suppressing the discontinuous patterns that appear around feature boundaries.

In addition to providing a global aggregation for the output $\textbf{z}$ at each sampling point, the linear Galerkin-type attention \eqref{eq:galerkin} has the ability to obtain a quasi-optimal approximation in the current approximation space spanned with the columns of $Q$~\cite{cao2021choose}. However, the expressive capability of the current bases in $Q$, $K$, and $V$ solely depends on the current input LR image through the latent representation $\textbf{z}$, is there any chance that we can enrich the bases which contains some extra and useful information for SR reconstruction, but not included in the input $\textbf{z}$? For this purpose, the point-wise FFN $\mathcal{O}: \mathbb{R}^{d_z}\rightarrow\mathbb{R}^{d_z}$ in Fig.\ref{fig:srno} (b) introduces nonlinearties on one hand, and the positions concatenated in $\textbf{z}$ enhance the bases on the other. In this way, the basis functions are being constantly enriched, and we reformulate the iterative process (See supplementary for a principled discussion):
 \begin{equation}
 z_{t+1}(x) = z_t(x)+\mathcal{O}((\mathcal{K}_t(z_t))(x) + z_t(x)). 
 \end{equation}

\textbf{Network details}. The Network architecture is shown in Fig.\ref{fig:srno}. As to the feature encoder $E_{\psi}$, we employ EDSR-baseline~\cite{lim2017enhanced}, or RDN~\cite{zhang2018residual}, both of which drop their upsampling layers, and their output channel dimensions $d_e=64$. The CNNs in $E_\psi$ assist SRNO in capturing the common basis functions from the ensemble of training samples, while the Galerkin-type attention layers provide the instance-specific basis enhancement. We employ the multi-head attention scheme in \cite{vaswani2017attention} by dividing the queries, keys and values into $n_{heads}$ parts with each of dimension $d_z/n_{heads}$. In our implementation, $d_z=256,n_{heads}=16$, yielding 16-dimensional output values. We only use two iterations $(T=2)$ in the kernel integral operator, which already outperforms previous works, while keeping the running time advantage. Note that we utilize $1\times1$ convolutions to replace all the linear layers in SRNO, since they have a GPU-friendly data structure. The detailed network structures are listed in the supplementary. 

\begin{figure*}
    \centering
    \tabcolsep=0.02in
    \begin{tabular}{cccccc}
        \raisebox{0.1in}{\rotatebox{90}{DIV2K($\times12$)}}
        \includegraphics[width=1.348in, height=1in]{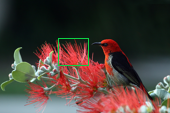}&
        \includegraphics[width=1in]{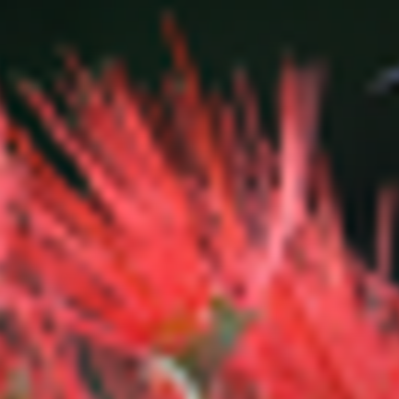} &
        \includegraphics[width=1in]{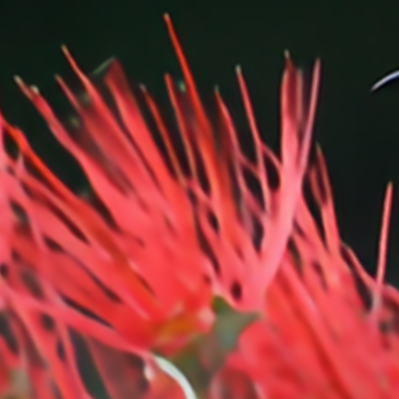} & \includegraphics[width=1in]{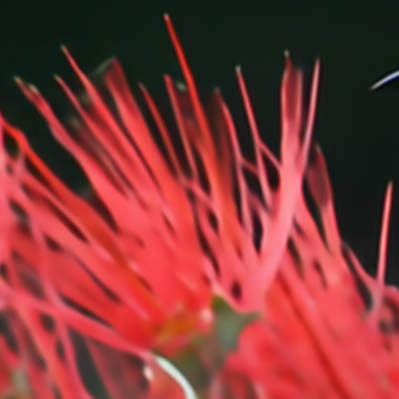} &
        \includegraphics[width=1in]{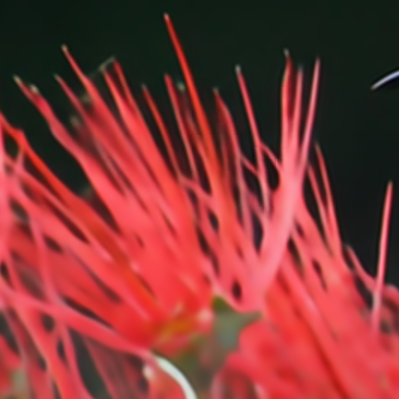} & \includegraphics[width=1in]{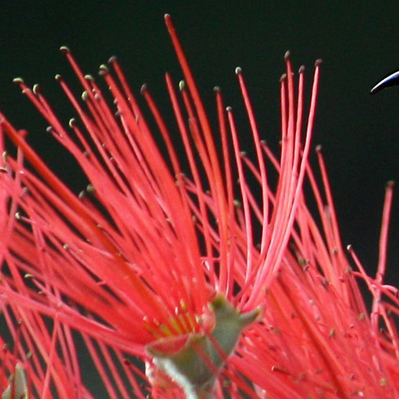} \\
        
        \raisebox{0.075in}{\rotatebox{90}{Urban100($\times8$)}}
        \includegraphics[width=1.348in, height=1in]{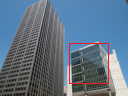}&
        \includegraphics[width=1in]{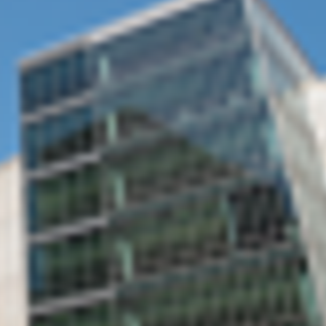} &
        \includegraphics[width=1in]{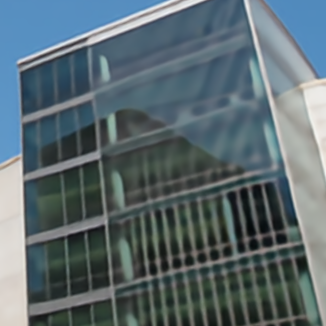} & \includegraphics[width=1in]{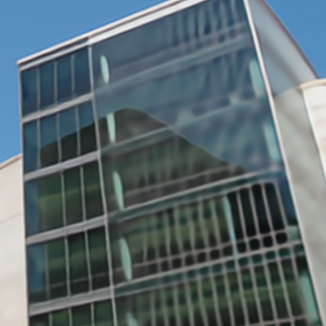} &
        \includegraphics[width=1in]{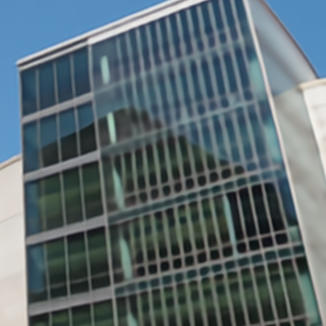} & \includegraphics[width=1in]{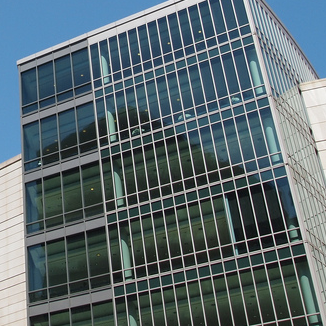} \\
        
        \begin{tabular}{c} LR Image \end{tabular} & Bicubic & LIIF~\cite{chen2021learning} & LTE~\cite{lee2022local} & SRNO (ours) & GT
    \end{tabular}
    \caption{\textbf{Visual comparison on other zero-shot SR}. The boxes in the first column indicate the areas that the close-ups on the right display. All methods are trained with continuous random scales in $\times1$--$\times4$ and tested on $\times8/\times12$ to evaluate the generalization capability to unknown scaling factors. RDN is used as the encoder for all methods.}
    %\vspace{-0.5em}
    \label{fig:qualitative}
\end{figure*}

\begin{figure*}
    \centering
    \tabcolsep=0.02in
    \begin{tabular}{ccccc}
        \raisebox{0.1in}{\rotatebox{90}{Urban100($\times6.6$)}}
        \includegraphics[width=1.27in, height=1.01in]{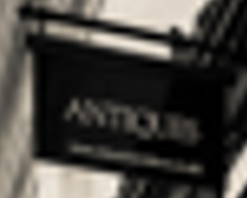}&
        \includegraphics[width=1.27in, height=1.01in]{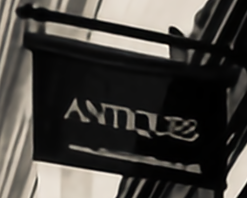} &
        \includegraphics[width=1.27in, height=1.01in]{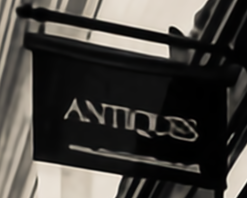} & \includegraphics[width=1.27in, height=1.01in]{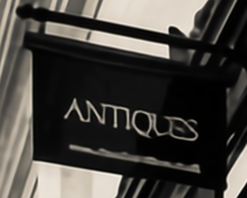} &
        \includegraphics[width=1.27in, height=1.01in]{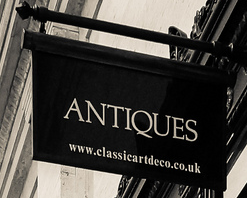}  \\
        
        \raisebox{0.075in}{\rotatebox{90}{Urban100($\times8.8$)}}
        \includegraphics[width=1.27in, height=1.01in]{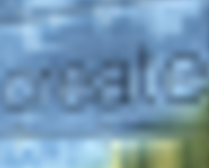}&
        \includegraphics[width=1.27in, height=1.01in]{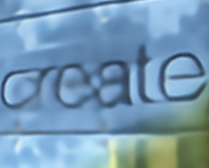} &
        \includegraphics[width=1.27in, height=1.01in]{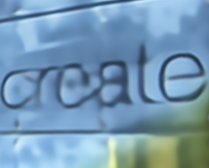} & \includegraphics[width=1.27in, height=1.01in]{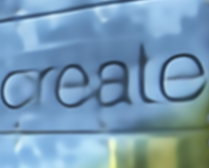} &
        \includegraphics[width=1.27in, height=1.01in]{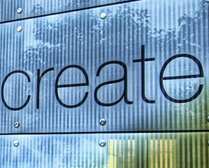} \\
        
        Bicubic & LIIF~\cite{chen2021learning} & LTE~\cite{lee2022local} & SRNO (ours) & GT
    \end{tabular}
    \caption{\textbf{Visual comparison on non-integer scales}. All methods use RDN as the encoder and are trained with continuous random scales in $\times1$--$\times4$.}
    %\vspace{-0.5em}
    \label{fig:non_int}
\end{figure*}

\begin{table*}
    \centering
    \scalebox{0.84}{
    \begin{tabular}{c|ccc|ccccc}
        \multirow{2}{*}{Method} & \multicolumn{3}{c|}{In-distribution} & \multicolumn{5}{c}{Out-of-distribution} \\
        & $\times$2 & $\times$3 & $\times$4 & $\times$6 & $\times$12 & $\times$18 & $\times$24 & $\times$30 \\
        \hline
        Bicubic & 31.01 & 28.22 & 26.66 & 24.82 & 22.27 & 21.00 & 20.19 & 19.59 \\
        EDSR-baseline~\cite{lim2017enhanced} & 34.55 & 30.90 & 28.94 & - & - & - & - & - \\
        EDSR-baseline-MetaSR~\cite{chen2021learning,hu2019meta} & 34.64 & 30.93 & 28.92 & 26.61 & 23.55 & 22.03 & 21.06 & 20.37 \\
        EDSR-baseline-LIIF~\cite{chen2021learning} & 34.67 & 30.96 & 29.00 &
        26.75 & 23.71 & 22.17 & 21.18 & 20.48 \\
        EDSR-baseline-LTE~\cite{lee2022local} & 34.72 & 31.02 & 29.04 &
        26.81 & 23.78 & 22.23 & 21.24 & 20.53 \\
        EDSR-baseline-SRNO (ours) & \textbf{34.85} & \textbf{31.11} & \textbf{29.16} &
        \textbf{26.90} & \textbf{23.84} & \textbf{22.29} & \textbf{21.27} & \textbf{20.56} \\
        \hline
        RDN-MetaSR~\cite{chen2021learning,hu2019meta} & 35.00 & 31.27 & 29.25 & 26.88 & 23.73 & 22.18 & 21.17 & 20.47 \\
        RDN-LIIF~\cite{chen2021learning} & 34.99 & 31.26 & 29.27 & 26.99 & 23.89 & 22.34 & 21.31 & 20.59 \\
        RDN-LTE~\cite{lee2022local} & 35.04 & 31.32 & 29.33 & 27.04 & 23.95 & 22.40 & 21.36 & 20.64 \\
        RDN-SRNO (ours) & \textbf{35.16} & \textbf{31.42} & \textbf{29.42} &
        \textbf{27.12} & \textbf{24.03} & \textbf{22.46} & \textbf{21.41} & \textbf{20.68} \\
    \end{tabular}}
    \caption{\textbf{Quantitative comparison on DIV2K validation set (PSNR (dB))}. The best performance are bolded. EDSR-baseline trains separate models for the three in-distribution scales. The rest methods use a single model for all scales, and are trained with continuous random scales uniformly sampled in $\times1$--$\times4$.}
    \label{tab:1}
    %\vspace{-0.5em}
\end{table*}

\begin{table*}
    \centering
    \scalebox{0.84}{
    \begin{tabular}{c|ccc|cc|ccc|cc}
        \multirow{3}{*}{Method} &  \multicolumn{3}{c|}{In-distribution} & \multicolumn{2}{c|}{Out-of-distribution} & \multicolumn{3}{c|}{In-distribution} & \multicolumn{2}{c}{Out-of-distribution}\\
        & $\times$2 & $\times$3 & $\times$4 & $\times$6 & $\times$8 & $\times$2 & $\times$3 & $\times$4 & $\times$6 & $\times$8 \\
        \cline{2-11}
        & \multicolumn{5}{c|}{Set5} & \multicolumn{5}{c}{Set14}\\
        \hline
        RDN~\cite{zhang2018residual} & 38.24 & 34.71 & 32.47 & - & - & 34.01 & 30.57 & 28.81 & - & -  \\
        RDN-MetaSR~\cite{hu2019meta,chen2021learning} & 38.22 & 34.63 & 32.38 & 29.04 & 26.96 & 33.98 & 30.54 & 28.78 & 26.51 & 24.97 \\
        RDN-LIIF~\cite{chen2021learning} & 38.17 & 34.68 & 32.50 & 29.15 & 27.14 & 33.97 & 30.53 & 28.80 & 26.64 & 25.15 \\
        RDN-LTE~\cite{lee2022local} & 38.23 & 34.72 & 32.61 & 29.32 & 27.26 & 34.09 & 30.58 & 28.88 & 26.71 & 25.16 \\
        RDN-SRNO (ours) & \textbf{38.32} & \textbf{34.84} & \textbf{32.69} & \textbf{29.38} & \textbf{27.28} & \textbf{34.27} & \textbf{30.71} & \textbf{28.97} & \textbf{26.76} & \textbf{25.26} \\
        \cline{2-11}
        & \multicolumn{5}{c|}{B100} & \multicolumn{5}{c}{Urban100}\\
        \cline{2-11}
        RDN~\cite{zhang2018residual} & 32.34 & 29.26 & 27.72 & - & - & 32.89 & 28.80 & 26.61 & - & -\\
        RDN-MetaSR~\cite{hu2019meta,chen2021learning} & 32.33 & 29.26 & 27.71 & 25.90 & 24.83 & 32.92 & 28.82 & 26.55 & 23.99 & 22.59\\
        RDN-LIIF~\cite{chen2021learning} & 32.32 & 29.26 & 27.74 & 25.98 & 24.91 & 32.87 & 28.82 & 26.68 & 24.20 & 22.79\\
        RDN-LTE~\cite{lee2022local} & 32.36 & 29.30 & 27.77 & 26.01 & 24.95 & 33.04 & 28.97 & 26.81 & 24.28 & 22.88 \\
        RDN-SRNO (ours) & \textbf{32.43} & \textbf{29.37} & \textbf{27.83} & \textbf{26.04} & \textbf{24.99} & \textbf{33.33} & \textbf{29.14} & \textbf{26.98} & \textbf{24.43} & \textbf{23.02}\\

    \end{tabular}}
    \caption{\textbf{Quantitative comparison on benchmark datasets (PSNR (dB))}. The best performances are in bold. RDN trains separate models for the three in-distribution scales. The rest methods use a single model for all scales, and are trained with continuous random scales uniformly sampled in $\times1$--$\times4$.}
    \label{tab:2}
\end{table*}

%-------------------------------------------------------------------------
\section{Experiments}
\subsection{Training}
\textbf{Dataset}. We used the DIV2K dataset~\cite{agustsson2017ntire} for network training, while the DIV2K validation set~\cite{agustsson2017ntire} and four benchmark datasets, including Set5~\cite{bevilacqua2012low}, Set14~\cite{zeyde2010single}, B100~\cite{martin2001database} and Urban100~\cite{huang2015single}, for evaluation. Peak signal-to-noise ratio (PSNR) is used as the evaluation metric. Similar to \cite{hu2019meta,chen2021learning}, we cropped image boundaries when computing PSNRs.

\textbf{Implementation Details}. Numerically, we have to explore the underlying continuous image function through access to its point-wise evaluations on the HR and LR grids. Let $B$ be the batch size and $n_{h_c}$ be the LR sampling counts. We first sample $B$ random scales $r^{(i)}$ following a uniform distribution $\mathcal{U}(1,4)$, and then crop $B$ patches of sizes $\{\sqrt{n}_{h_c}r^{(i)}\times\sqrt{n}_{h_c}r^{(i)}\}_{i=1}^B$ from the HR training images (one per each). We find $n_{h_c} = 128^2$ is sufficient for SRNO to represent most natural image patches and fix this choice in the sequel. The LR counterparts are downsampled using bicubic interpolation with the corresponding $r^{(i)}$. In order to keep a consistent number of supervision points for the LR patches of different scales in a single batch, we randomly sample $128^2$ HR pixels and calculate the corresponding fractional coordinates on the coarse grid associated with $r^{(i)}$, as is done in \cite{chen2021learning}. We use an L1 loss~\cite{lim2017enhanced} and the Adam~\cite{kingma2015adam} optimizer with an initial learning rate $4\times 10^{-5}$ and the maximum $4\times 10^{-4}$. All models are trained for 1000 epochs with batch size 64, and the learning rate decays by the cosine annealing after a warm-up phase of 50 epochs. 

\subsection{Evaluation}
\textbf{Quantitative result}. On the DIV2K validation set, Table \ref{tab:1} presents a quantitative comparison among our SRNO and three existing arbitrary-scale SR methods, MetaSR~\cite{hu2019meta,chen2021learning}, LIIF~\cite{chen2021learning}, and LTE~\cite{lee2022local}. Results when EDSR-baseline~\cite{lim2017enhanced}, and RDN~\cite{zhang2018residual} are used as encoders are displayed in the top and bottom rows, respectively. One can observe that SRNO provides the best results over all scale factors, irrespective of the encoder employed. Additionally, on in-distribution scales ($\times2,\times3,\times4$), our method outperforms earlier works by wide margins. 

As RDN facilitates better reconstruction accuracy than EDSR-baseline in Tab.\ref{tab:1}, we choose RDN as the encoder in the comparisons over the four benchmark datasets. The results are listed in Tab.\ref{tab:2}, where SRNO obtains all the best performances. Note that only SRNO produces substantial improvements over RDN on every in-distribution scale, although the latter trains separate models for each scale. As the experimental settings in Tab.\ref{tab:1} and \ref{tab:2} only differ in the decoder part, we conclude that the Galerkin-type attention mechanism employed in SRNO does contribute to better function approximation capability.

\textbf{Qualitative result}. Figure \ref{fig:qualitative} illustrates the visual results obtained from two images in the Urban100 and DIV2K validation datasets. Both the regions marked with a rectangle are rich in high-frequency structures. The bicubic interpolations exhibit severe aliasing artifacts, which obviously have been transfered into the latent codes of LIIF~\cite{chen2021learning} as it directly relies on the interpolation of the feature maps and the bilinear ensemble. Thanks to the local Fourier frequency estimation step, LTE~\cite{lee2022local} performs somewhat better in partial regions (e.g. the top-right corner of the building in the second row), but aliasing still dominates on the whole. In contrast, SRNO successfully recovers the fine structures, which again verifies the importance of the global integral kernels in capturing the correct overall structures.

Figure \ref{fig:non_int} shows two examples of scene text images. In the first row, only SRNO can clearly recover the word ”ANTIQUES”. And in the second row, only SRNO can consistently restore the two e's. Obviously, the Galerkin-type attention attributes to this global consistency.

\subsection{Ablation Study}

\begin{figure}
\centering
\includegraphics[scale=0.74]{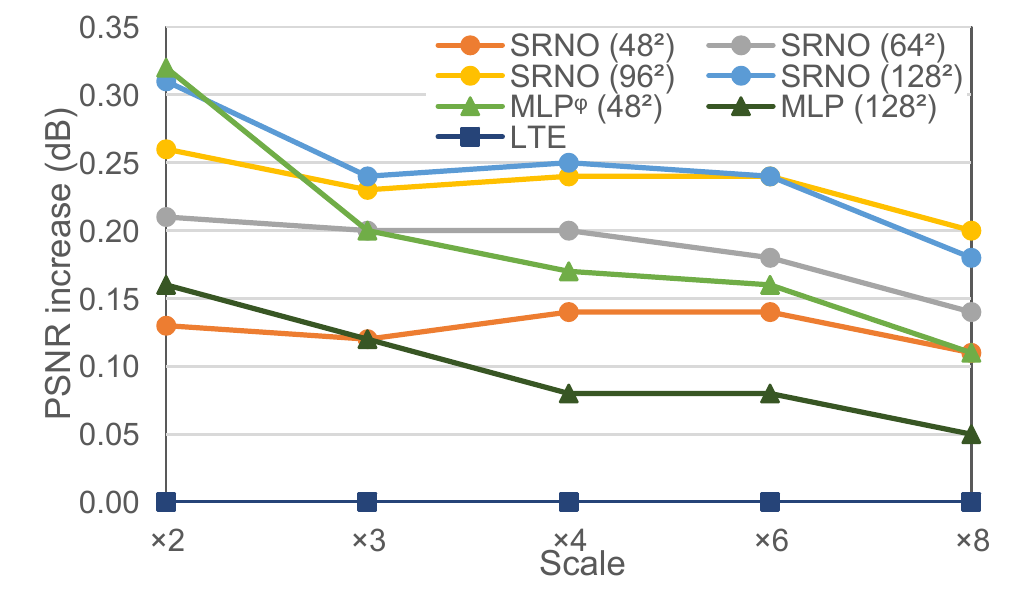}
\caption{\textbf{Data samples vs. PSNRs}. Evalutated on the Urban100. We train models using $48^2/64^2/96^2/128^2$ pixel samples and calculate the PSNR increment relative to LTE. $^\varphi$ indicates training on DIV2K~\cite{agustsson2017ntire} (800 images) plus Flickr2K~\cite{timofte2017ntire} (2650 images) dataset.}
\vspace{-0.5em}
\label{fig:sample}
\end{figure}

\begin{figure}
    \centering
    \tabcolsep=0.02in
    \begin{tabular}{ccc}
        \includegraphics[width=1.02in]{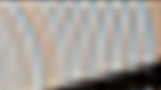}&
        \includegraphics[width=1.02in]{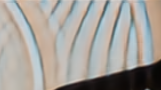}&
        \includegraphics[width=1.02in]{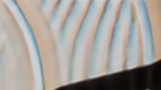} \\
       Bicubic & SRNO ($48^2$) & SRNO ($64^2$)\\
        
        \includegraphics[width=1.02in]{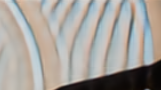}&
        \includegraphics[width=1.02in]{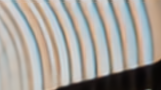}&
        \includegraphics[width=1.02in]{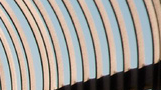} \\
       SRNO ($96^2$) & SRNO ($128^2$) & GT
    \end{tabular}
    \caption{\textbf{Effects of sampling global image structures}. Test on $\times 8$ scale. EDSR-baseline is used as the encoder, while the data in parentheses indicate the pixel numbers sampled in the HR images during trainning.}
    %\vspace{-0.5em}
    \label{fig:sample2}
\end{figure}

\textbf{Data sampling}. The quantity of sampling points is crucial in our super-resolution operator learning architecture. Figure \ref{fig:sample} shows that almost all scales perform better when using additional pixel samples. However, the benefit becomes less as the numbers of samples increases. $96^2$ or $128^2$ points would be good trade-offs between training time and performance. Figure \ref{fig:sample2} shows that more sampling points can more accurately represent an RGB-based function, which facilitates to learn the map between two approximation spaces. Using the DIV2K (800 images) and Flickr2K (2650 images) datasets, we also train an EDSR-baseline-MLP with position embedding, denoted by MLP$^\varphi$. We can see that our SRNO ($64^2/96^2/128^2$) obtains better performance for scales $\times 3,\times4,\times6,\times8$ , trained with only $800$ images in DIV2K. The curve for MLP($128^2$) in Fig.\ref{fig:sample} reflects the performance advantage of SRNO($128^2$), both trained with DIV2K dataset. Random and sequential sampling results are presented in Supplementary.

\begin{table}
    \centering
    \scalebox{0.828}{
    \begin{tabular}{c|c|c}
        {Method} & {\#Params. (M)} & {\#FLOPs (G)} \\
        \hline
        EDSR-baseline-LIIF & 1.6 & 85.0\\
        EDSR-baseline-LTE & 1.7 & 75.3\\
        EDSR-baseline-SRNO (ours) & 2.0 & 65.8\\
        \hline
        RDN-LIIF & 22.4 & 765.2\\
        RDN-LTE & 22.5 & 755.5\\
        RDN-SRNO (ours) & 22.8 & 746.0\\
    \end{tabular}}
    \caption{\textbf{Comparison of model parameters and FLOPs}. Metrics are measured under the setting of $128^2$ sample points.}
    \label{tab:flops}
    %\vspace{-5pt}
\end{table}

\begin{table}
    \centering
    \scalebox{0.828}{
    \begin{tabular}{c|ccc|cc|c}
        \multirow{2}{*}{Method} & \multicolumn{3}{c|}{In-dis.} & \multicolumn{2}{c}{Out-of-dis.} & \multirow{2}{*}{Time (ms)}\\
        & $\times$2 & $\times$3 & $\times$4 & $\times$6 & $\times$8 \\
        \hline
        SRNO & 33.83 & \textbf{30.50} & \textbf{28.79} & 26.55 & \textbf{25.05} & 149 \\
        SRNO (-i) & 33.81 & 30.47 & 28.76 & 26.55 & 25.00 & 156 \\
        SRNO (-h) & \textbf{33.87} & 30.47 & 28.76 & \textbf{26.56} & 25.02 & 153 \\
        SRNO (-w) & 33.80 & 30.45 & 28.75 & 26.53 & 25.00 & 124 \\
        SRNO (-l) & 33.82 & 30.45 & 28.75 & 26.55 & 25.00 & \textbf{117} \\
        \hline
        LIIF~\cite{chen2021learning} & 33.68 & 30.36 & 28.64 & 26.46 & 24.94 & 183 \\
        LTE~\cite{lee2022local} & 33.72 & 30.37 & 28.65 & 26.50 & 24.99 & 205 \\
        LTE+~\cite{lee2022local} & 33.71 & 30.41 & 28.67 & 26.49 & 24.98 & 160 \\
    \end{tabular}}
    \caption{\textbf{Quantitative ablation study on design choices of SRNO}. Evalutated on the Set14 (PSNR(dB)). -h/w/l refers to $n_{heads}=8$, $d_z=128$, and $T=1$ correspondingly. -i refers to using nearest-neighborhood interpolation and position embedding. EDSR-baseline is used in lifting. The running time is averaged over Set14 on $\times2/3/4/6/8$ SR. For each scale SR, we run every model three times and choose the minimum running time.}
    \label{tab:ab2}
\end{table}

\begin{figure}
    \centering
    \tabcolsep=0.02in
    \begin{tabular}{cc}
        \includegraphics[width=1.5in]{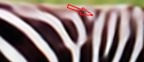}&
        \includegraphics[width=1.5in]{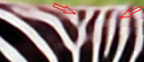} \\
        LTE & SRNO (-i)\\
        
        \includegraphics[width=1.5in]{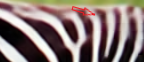}&
        \includegraphics[width=1.5in]{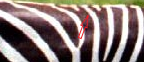} \\
        SRNO & GT
    \end{tabular}
    \caption{\textbf{Interpolation vs. interpolation-free of SRNO}. Test on $\times 8$ scale. EDSR-baseline is used as an encoder. -i refers to using nearest-neighborhood interpolation and position embedding.}
    %\vspace{-0.5em}
    \label{fig:ab}
\end{figure}

\textbf{Other design choices}. In Tab.\ref{tab:ab2}, we retrain the following models using EDSR-baseline~\cite{lim2017enhanced}. All the tests were conducted using a single NVIDIA RTX 3090. In comparing our interpolation-free method to SRNO(-i) using nearest-neighbor interpolation, we find that it works better and faster. It should be noted that we only generate the image once, as opposed to employing an MLP four times as in LIIF~\cite{chen2021learning} and LTE~\cite{lee2022local}. Figure \ref{fig:ab} shows that SRNO, compared to SRNO(-i), alleviates the blocky artifact, and compared to the local ensemble~\cite{lee2022local}, overcomes the over-smoothing problem. By comparing SRNO with SRNO(-h), it can be shown that cutting the number of heads from $16$ to $8$ causes an increase in time costs because GPU's parallelism is reduced. We see that reducing the number of basis functions and the iterative updating layers results in a considerable performance drop by comparing SRNO with SRNO(-w) and SRNO(-l). The number of basis functions ($d_z$) and the number of subspaces ($n_{heads}$) play an important role in the approximation power of our model.

%\begin{figure}
%    \centering
%    \tabcolsep=0.02in
%    \begin{tabular}{ccc}
%        \multirow{2}{*}{\includegraphics[width=0.9in]{cvpr2023/figure/set14_7_gt.png}}&
%        \includegraphics[width=0.9in]{cvpr2023/figure/0_1.png}&
%        \includegraphics[width=0.9in]{cvpr2023/figure/1_1.png}\\
%        
%        &
%        \includegraphics[width=0.9in]{cvpr2023/figure/0_8.png}&
%        \includegraphics[width=0.9in]{cvpr2023/figure/1_8.png}\\
%        
%        GT & t = 1 & t = 2
%    \end{tabular}
%    \caption{\textbf{Dynamic basis update}. The annotation $t=k$ refers to the iterative layer number. The pictures display that the learned basis functions, dynamically updated in the approximation subspaces.}
    %\vspace{-0.5em}
%    \label{fig:basis}
%\end{figure}
\begin{figure}
\centering
\includegraphics[scale = 0.35]{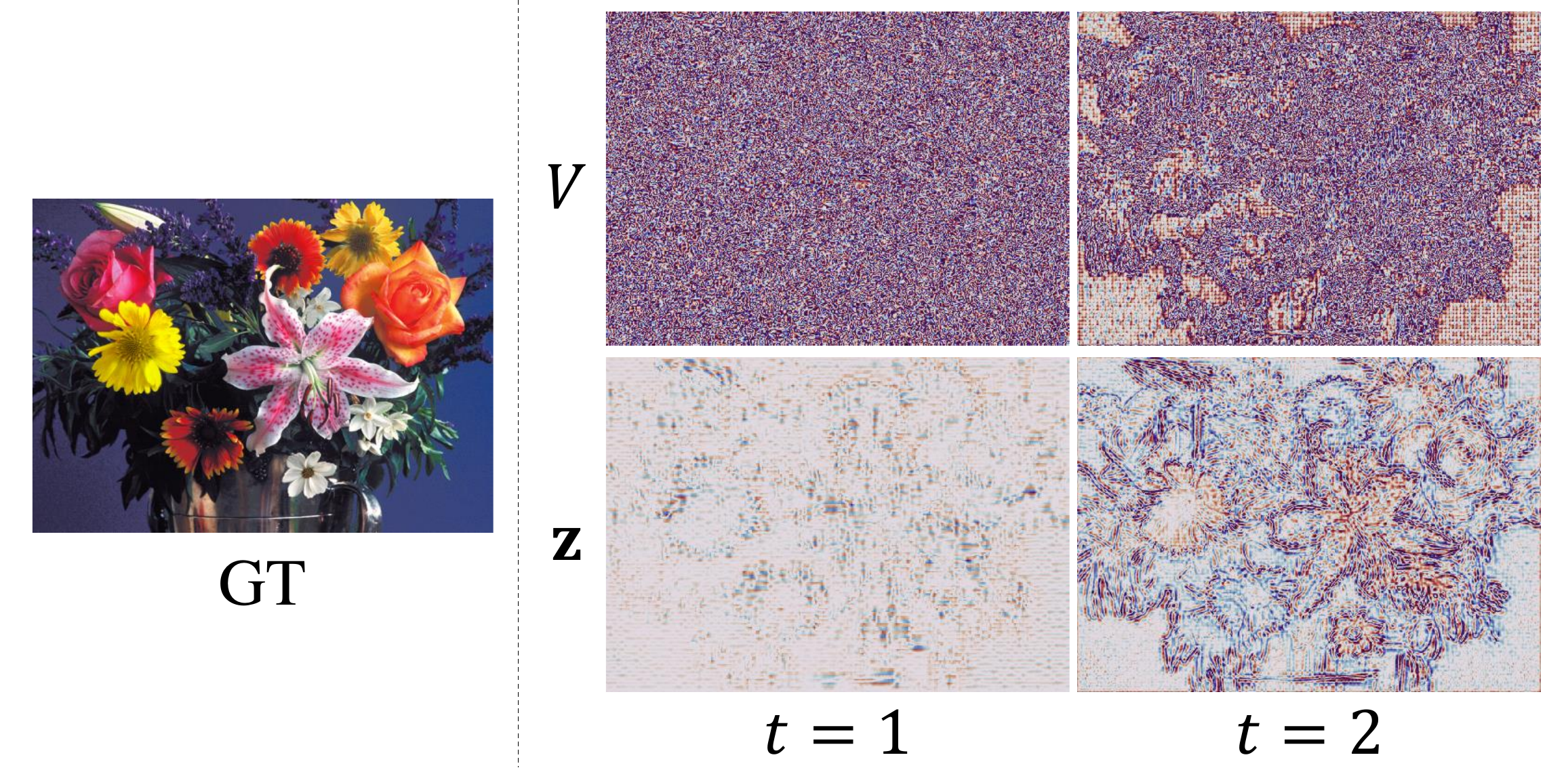}
\caption{\textbf{Dynamic basis update}. The annotation $t=k$ refers to the iterative layer number. The pictures display one of the learned basis functions in $V$ and the latent representations $\textbf{z}$}
\label{fig:basis}
\end{figure}

\begin{figure}
\centering
\includegraphics[scale = 0.73]{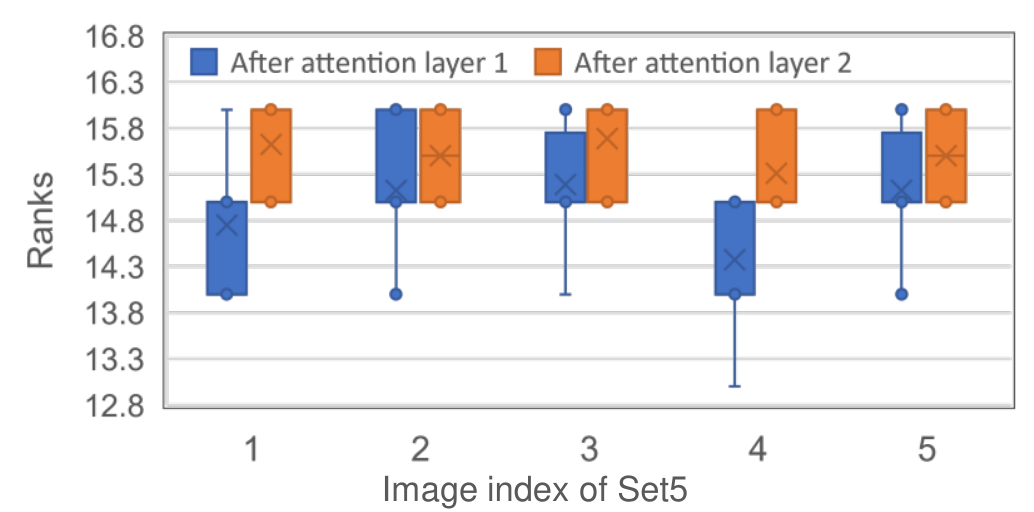}
\caption{\textbf{Ranks of the matrix $\textbf{z}$}. Evaluated on the Set5 dateset.}
\label{fig:rank}
\end{figure}

\subsection{Dynamic Basis} 
We investigate the dynamic basis updating phenomenon across the approximation subspaces in this section. We show two exemplar basis functions in the two consecutive Galerkin attention layers in Fig.\ref{fig:basis}. The basis function in the second layer appears to be more structured than in the previous layer. The dynamic update of the approximation subspaces spanned by the column vectors of $Q/K/V$ comes from two factors, the concatenated random coordinates and the nonlinearity introduced by the FNN. Without them, the Galerkin-type attention would just be a linear combination of the bases in the current approximation subspaces, and SRNO would not be able to take any chances to enrich the basis for every individual image through optimizations. The increasing ranks of the latent representation matrix $\textbf{z}$, shown in Fig.\ref{fig:rank}, provide the evidence that Galerkin-type attention is becoming more complex than a straightforward linear combination of the existing bases. Due to the space limitation, we have provided a principled discussion on the dynamic basis update processes in the supplementary material.

\section{Conclusion}
In this paper, we proposed the Super-Resolution Neural Operator (SRNO) for continuous super-resolution. In SRNO, each image is seen as a function, and our method learns a map between finite-dimensional function spaces, which allows SRNO can be trained and generalize on different levels of discretization. First, in the Lifting, we use CNN-based encoders to capture feature maps from LR images and design a simple but efficient interpolation-free method, addressing the discretization inconsistency problem encountered in SR. Second, to approximate our target function in the Iterative Kernel Integration, we employ a linear attention operator that has been proved to be comparable to the Petrov-Galerkin projection. Finally, we map the last hidden representation to the output function. Experimental results show that our SRNO outperforms other arbitrary-scale SR methods in performance as well as computation time, and particularly in the capability of capturing the global image structures.

%%%%%%%%% REFERENCES

\clearpage
\twocolumn[{%
\maketitle
\hsize=\textwidth
\centering
 \Large \textbf{Supplemental Material:\\ Super-Resolution Neural Operator}
 \vspace{60pt}
}]

\begin{appendices}
\setcounter{equation}{0}
\setcounter{figure}{0}
\setcounter{table}{0}

\begin{figure*}
\centering
\includegraphics[scale = 0.65]{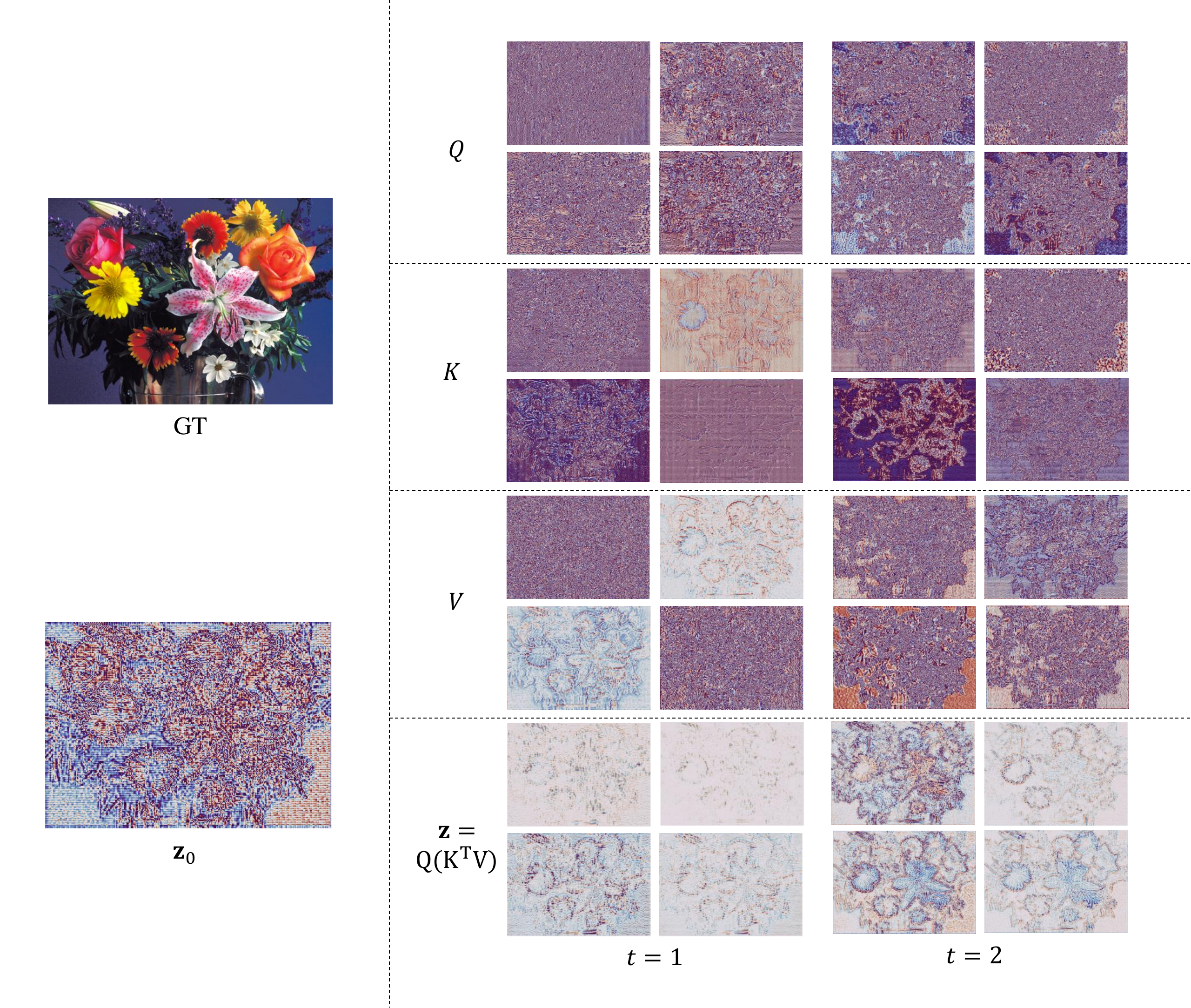}
\caption{\textbf{Dynamic basis update}. The annotation $t=k$ refers to the iterative layer number. We display four basis function evaluation vectors (columns) of the matrices $Q$, $K$, and $V$, respectively, as well as the synthesized  the latent representations $\textbf{z}$}
\label{fig:1}
\end{figure*}
This document provides additional details and results.

%%%%%%%%% BODY TEXT
\section{Dynamic basis update}

%\textbf{Attention mechanism and variants}.
%The attention mechanism is the key component of Transformer~\cite{vaswani2017attention}, which can be used to explore the underlying corelations among tokens and to adaptively aggregate valuable information. However, the computation complexity in vanilla Transformer, due to the softmax function, is quadratic in the signal length. Recent works of linearizing attention rely on the assumption of the existence of methods to approximate the softmax kernel~\cite{katharopoulos2020transformers,choromanski2020rethinking,peng2021random,qin2021cosformer,wu2022flowformer}. Another type of linearization exploits the low-rank nature of the matrix product using various methods such as sampling or projection~\cite{blanc2018adaptive,rawat2019sampled,song2021implicit,wang2020linformer}, or fast multipole decomposition~\cite{nguyen2021fmmformer}. However, these two kinds of methods still focus on the competition between similarity weights that softmax introduced. Galerkin-type attention~\cite{cao2021choose}, inspired by the conjecture in \cite{schlag2021linear}, removed the softmax overall and achieved impressive results in PDEs. The Nyström approximation~\cite{xiong2021nystromformer} and the kernel interpretation~\cite{tsai2019transformer} acknowledges the similarity between the attention matrix and an integral kernel defined in neural operators~\cite{kovachki2021neural}.

In this section, we provide a principled discussion on the quasi-optimal property of the Galerkin-type attention mechanism, which tells that in each attention layer, SRNO can achieve the approximation capability that the Petrov-Galerkin projection can offer. Although some theorems have been proved in \cite{cao2021choose2}, we provide a systematic and complete discussion on the dynamic basis update processes in the query, test and value approximation spaces. The background knowledge about the Galerkin projection can be found in \cite{ern2004theory2}

Let $g_\theta(\cdot): \mathbb{R}^{n\times d}\rightarrow\mathbb{Q}_h, \{z_t: D_t \rightarrow \mathbb{R}^d\}\mapsto \{z_{t+1}:D_{t+1} \rightarrow \mathbb{R}^d\}$ be a learnable map that is the composition of the Galerkin-type attention operator and the FFN $\mathcal{O}$ used in SRNO, where $\mathbb{Q}_h \subset \mathcal{H}(\Omega_{h})\subset\mathcal{H}$ is the current value space spanned by the column vectors of $Q=\mathbf{z}W_q$. Similarly, we can define the approximation spaces $\mathbb{V}_h$ and $\mathbb{K}_h$. Suppose $f_h$ is the best approximation of image function $f$ in current value space $\mathbb{Q}_h$, \textit{i.e.}, $f_h = \arg\min_{q\in \mathbb{Q}_h}\lVert f-q\rVert_\mathcal{H}$, and then the approximation error between $g_\theta(\cdot)$ and $f$ can be described as:
\begin{equation}
\label{eq:1}
    \lVert f-g_\theta(\mathbf{z})\rVert_\mathcal{H}\leq\lVert f_h-g_\theta(\mathbf{z})\rVert_\mathcal{H} + \lVert f-f_h\rVert_\mathcal{H}.
\end{equation}
Let $B(\cdot, \cdot): \mathbb{V}_h \times \mathbb{Q}_h \rightarrow \mathbb{R}$ be a continuous bilinear form. Here we define, for $(u,v)\in \mathbb{Q}_h \times \mathbb{V}_h$,  $B(u,v):= h^2\sum^n_{i=1}u(x_i)v(y_i)$. Under the settings of Galerkin-type attention~\cite{cao2021choose2}, for any given $q\in \mathbb{Q}_h$, we have 
\begin{equation}
    \max_{v\in\mathbb{V}_h}{\frac{|B(v,q)|}{\lVert v \rVert_\mathcal{H}}} \ge c\lVert q\rVert_\mathcal{H},
\end{equation}
\textit{i.e.}, $B$ is coercive on the current key space $\mathbb{V}_h$ with constant $c$. \eqref{eq:1} can be reformulated as 
\begin{equation}
\label{eq:2}
    \lVert f-g_\theta(\mathbf{z})\rVert_\mathcal{H}\leq c^{-1}\max_{v\in\mathbb{V}_h}{\frac{|B(v,f_h - g_\theta(\mathbf{z}))|}{\lVert v \rVert_\mathcal{H}}} + \lVert f-f_h\rVert_\mathcal{H}.
\end{equation}
Our purpose is to minimize \eqref{eq:2} by optimizing the trainable parameters $\theta$,
\begin{equation}
\label{eq:3}
    \min_\theta \max_{v\in\mathbb{V}_h}{\frac{|B(v,f_h - g_\theta(\mathbf{z}))|}{\lVert v \rVert_\mathcal{H}}} \leq \min_{q\in\mathbb{Q}_h} \max_{v\in\mathbb{V}_h}{\frac{|B(v,f_h - q)|}{\lVert v \rVert_\mathcal{H}}}.
\end{equation}
\eqref{eq:2} and \eqref{eq:3} show the approximation capacity of a Galerkin-type attention used in SRNO as the kernel integral operator. In SRNO, we are actually optimizing the basis functions of the current value space $\mathbb{Q}_h$ to approximate the best $f_h$. By the Riesz representation theorem~\cite{ciarlet2013linear2}, there exists a value-to-key linear map $\Pi:\mathbb{Q}_h\rightarrow\mathbb{V}_h$ such that $B(v,f_h)=\langle v,\Pi f_h\rangle$. In order to reveal the interactions among the bases of the three approximation spaces, we introduce the second bilinear form $A(\cdot, \cdot): \mathbb{K}_h\times \mathbb{V}_h\rightarrow\mathbb{R}$ to substitute the inner product $\langle v, \Pi f_h\rangle$. In practice, the FFN $\mathcal{O}$, as a universal approximator in $g_\theta$, helps the bilinear form $A(\cdot,\cdot)$ to approximate the inner product $\langle v, \Pi f_h\rangle$. We thus define the following problem to approximate the right hand side of \eqref{eq:3} ($j=1,\dots,d$):
\begin{equation}
\label{eq:4}
    \min_{q\in\mathbb{Q}_h}\max_{v\in\mathbb{V}_h}\frac{\lVert A(k_j,v)-B(q,v)\rVert}{\lVert v\rVert_\mathcal{H}},
\end{equation}
which involves solving the following operator equation system (finding $z_j\in \mathbb{Q}_h$ and $w\in \mathbb{V}_h$):
\begin{equation}
    \begin{split}
        \langle w,v\rangle + B(v,z_j) &= A(k_j,v), \quad \forall v\in \mathbb{V}_h,\\
        B(w,q)&=0, \quad \forall q\in\mathbb{Q}_h,
    \end{split}
\end{equation}
which is further equivalent to solve the following linear system:
\begin{equation}
    \begin{pmatrix} M & B^T \\ B & 0 \end{pmatrix} \begin{pmatrix} \boldsymbol{\mu}\\ \boldsymbol{\lambda} \end{pmatrix} = \begin{pmatrix} (V^TK)_j\\0 \end{pmatrix},
\end{equation}
where $\mathbb{Q}_h$, $\mathbb{V}_h$ is formd by sets of basis $\{q_j(\cdot)\}_{j=1}^r$, $\{v_j(\cdot)\}_{j=1}^d$, respectively. $B\in\mathbb{R}^{r\times d}$, $M\in\mathbb{R}^{d\times d}$, and $(V^TK)_j\in\mathbb{R}^{d\times 1}$. $ \boldsymbol{\mu}:=\mu(w)=(\mu_{v_1}(w),\dots,\mu_{v_d}(w))^T$ is the vector representation for $w(\cdot) =\sum_{j=1}^d\mu_{v_j}(w)v_j(\cdot)$. Similiar to $\boldsymbol{\lambda}$ for $z_j$. It is straightforward to verify that $h^2(V^TK)_{ij}=A(k_j,v_i)$, $B_{ij}=B(v_j,q_i)$, $B=h^2(QU)^TV, M_{ij}=\langle v_i, v_j\rangle$. And then we can get:
\begin{equation}
    \boldsymbol{\lambda}=(BM^{-1}B^T)^{-1}BM^{-1}(V^TK)_j,
\end{equation}
if $rank(Q)=r\leq rank(V)=d$, which is verified by our experiments. We multiply a  permutation matrix $U\in\mathbb{R}^{d\times d}$ to Q, such that $QU$'s first $r$ columns form the value vector $(q_j(x_1),\dots,q_j(x_n))^T$ as the bases $\{q_j(\cdot)\}_{j=1}^r$ of $\mathbb{Q}_h$. Then we multiply the permuted basis matrix $QU$ with $\begin{pmatrix}\boldsymbol{\lambda} & 0 \end{pmatrix} \in \mathbb{R}^d$, yielding
\begin{equation}
\begin{split}
    z_j &= h^2(QU)W(V^TK)_j,\\
    W &= \Lambda \begin{pmatrix}B \\ 0 \end{pmatrix} M^{-1},
\end{split}
\end{equation}
where $\Lambda = diag((BM^{-1}B^T),0)$. The layer normalization scheme for $V,K$ is used to mimick the matrix $W$.

The dynamic basis update rule, minimizer to \eqref{eq:4}, can be defined as:
\begin{equation}
    z_j(\cdot):=\sum^d_{l=1}A(\widetilde{k}_j,\widetilde{v}_l)q_l(\cdot),\quad j=1,\dots,d,
\end{equation}
where $\{\widetilde{k}_j\},\{\widetilde{v}_l\}$ are the column vectors of layer normnalized matrices $\widetilde{K},\widetilde{V}$.

The point-wise FFN $\mathcal{O}: \mathbb{R}^{d_z}\rightarrow\mathbb{R}^{d_z}$ introduces nonlinearties on one hand, and the positions concatenated in $\textbf{z}$ enhance the bases on the other. In this way, the basis functions not only approximate the functions in the current value space, but also are being constantly enriched. Note that, in practice,  we swap the matrix $K,V$ to make this process closer to self-attention in \cite{vaswani2017attention2}. In summary, our iterative process consist of two step: 1) the linear attention $Q(\widetilde{K}^T\widetilde{V})$ minimizes the \eqref{eq:4} in the current value space; 2) the point-wise FFN $\mathcal{O}$ and position information for the latent representation $\mathbf{z}$ enrich the basis functions. The dynamic basis updating phenomenon is demonstrated in Fig.\ref{fig:1}. We observe that the basis function in the second layer appears to be more structured than in the previous layer, which verifies the validity of our method.

\begin{figure}[b]
\centering
\includegraphics[scale = 0.41]{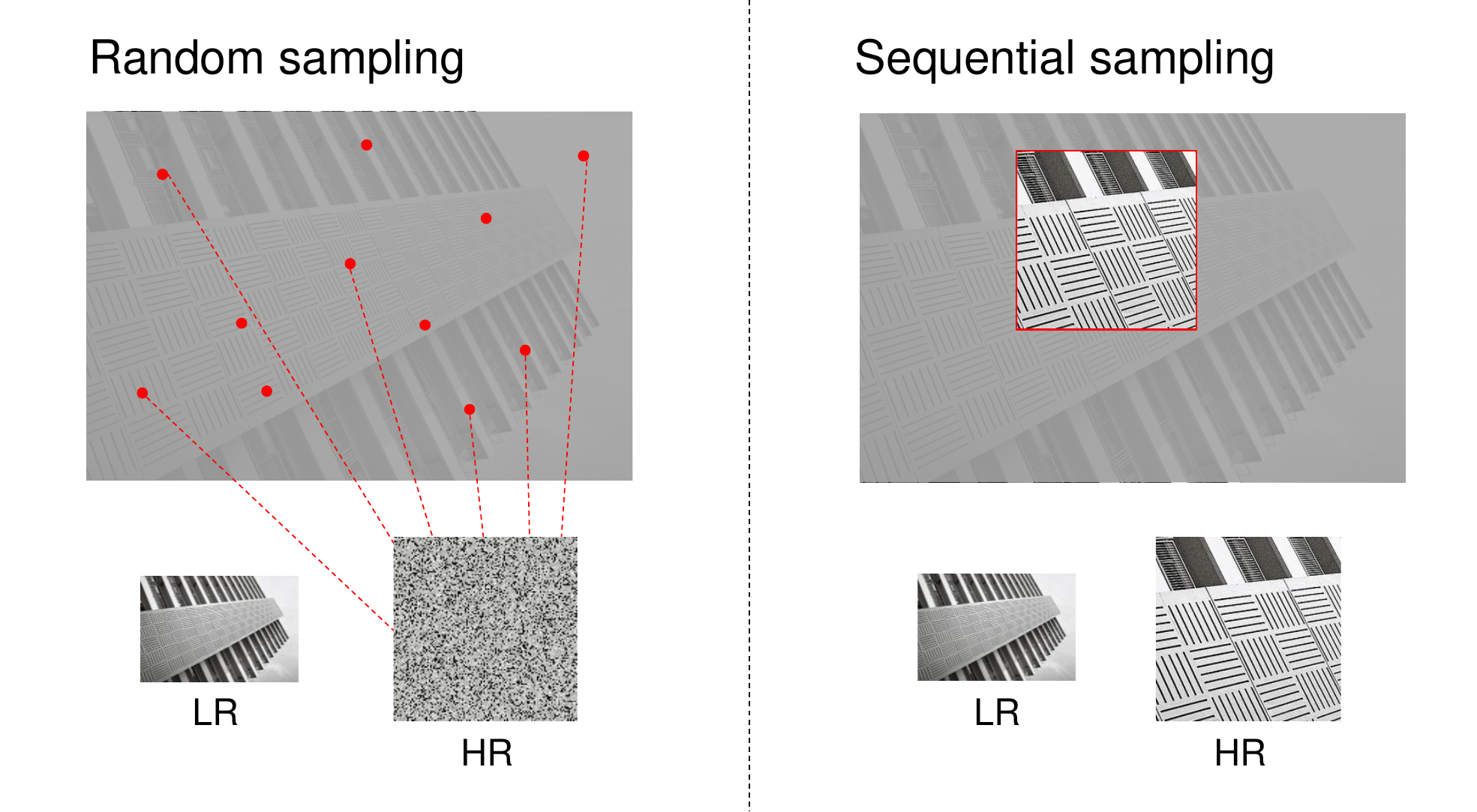}
\caption{\textbf{Random vs. Sequential sampling}.}
\label{fig:samplem}
\end{figure}

\begin{figure}
    \centering
    \tabcolsep=0.02in
    \begin{tabular}{ccc}
        \includegraphics[width=1.3in]{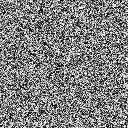} & \includegraphics[width=1.3in]{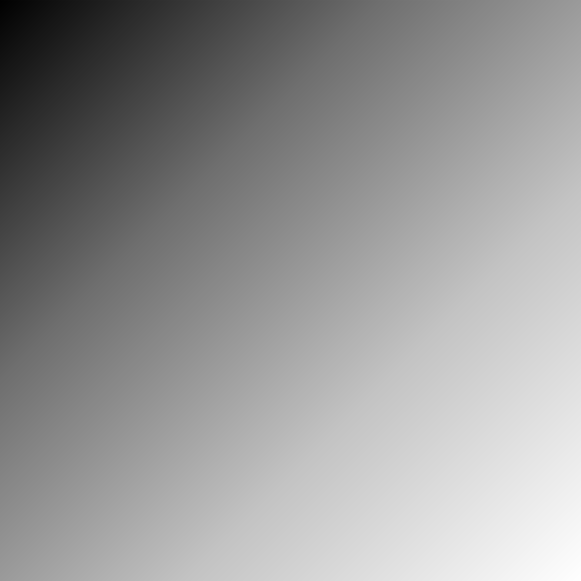} &
        \includegraphics[height=1.3in,width=0.1in]{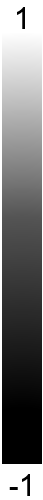}\\
        random coordinates & sequential coordinates
    \end{tabular}
    \caption{\textbf{Random v.s. sequential coordinates}.}
    %\vspace{-0.5em}
    \label{fig:coord}
\end{figure}

\begin{figure}
    \centering
    \tabcolsep=0.02in
    \begin{tabular}{ccc}
        %\raisebox{0.1in}{\rotatebox{90}{Urban100($\times4$)}}
        \includegraphics[width=1in]{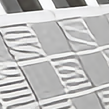} &
        \includegraphics[width=1in]{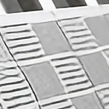} & \includegraphics[width=1in]{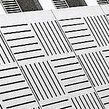} \\
        
         SRNO (-s) & SRNO & GT
    \end{tabular}
    \caption{\textbf{Visual comparison on sampling methods}. Test on $\times 4$ scale. EDSR-baseline is used as the encoder.}
    %\vspace{-0.5em}
    \label{fig:vs}
\end{figure}

\section{Network Architecture}

\begin{figure*}
\centering
\includegraphics[scale = 0.5]{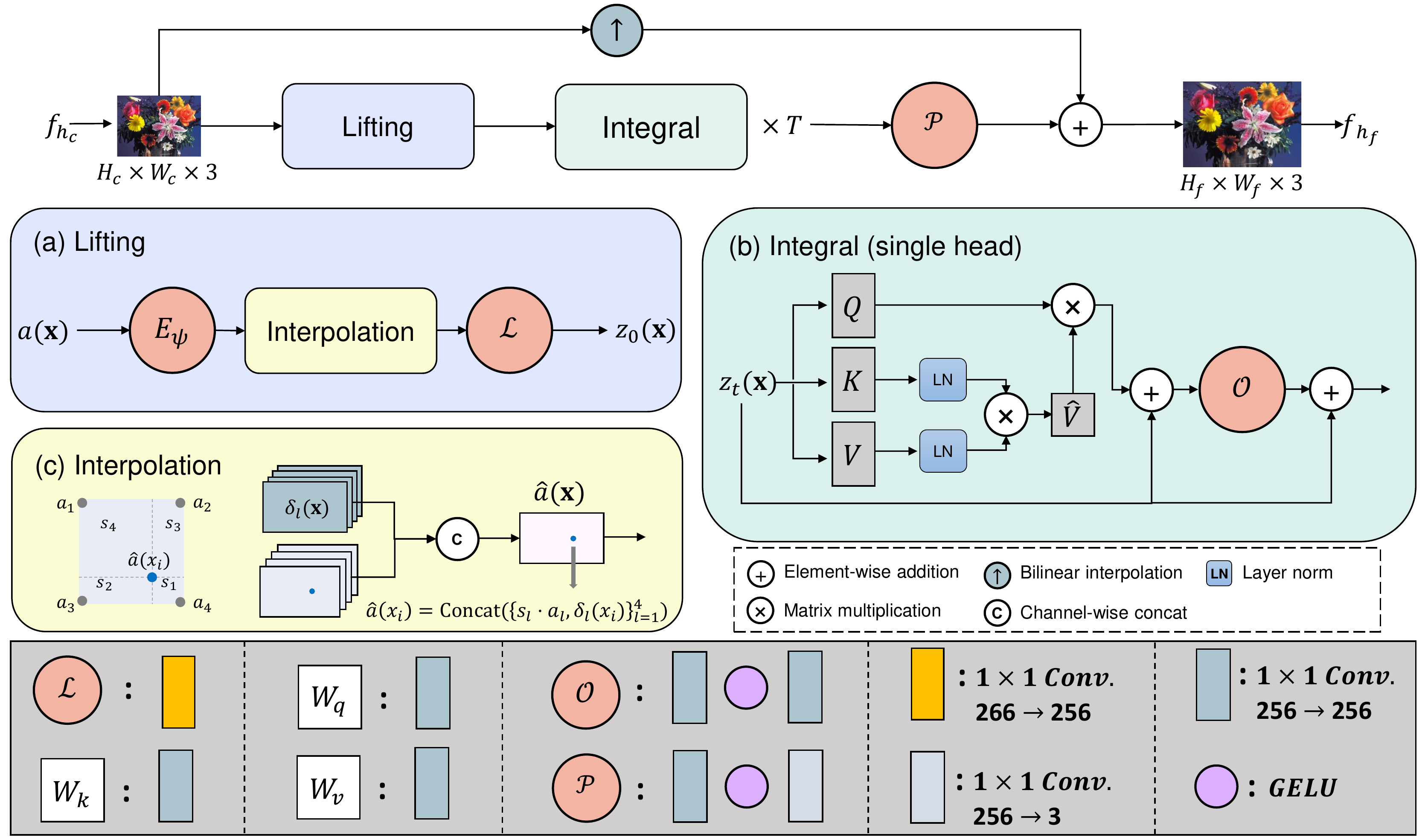}
\caption{\textbf{Super-resolution neural operator (SRNO) architecture for continuous SR.} Encoder $E_\psi$'s structure is omited.}
\label{fig:srnod}
\end{figure*}

The Network architecture is shown in Fig.\ref{fig:srnod}. The input LR image $f_{h_c}$ undergoes three phases to output the HR image $f_{h_f}$ with the specified resolution: (a) Lifting the LR pixel values $a(\textbf{x})$ on the set of coordinates $\textbf{x}=\{x_i\}_{i=1}^{n_{h_f}}$ to a higher dimensional feature space by a CNN-based encoder $E_\psi$, constructing the latent representation $\hat{a}(x)$, and linearly transforming into the first layer's input $z_0(\textbf{x})$. (b) kernel integrals composed of $T$ layers of Galerkin-type attention, and (c) finally project to the RGB space. As to the feature encoder $E_{\psi}$, we employ EDSR-baseline~\cite{lim2017enhanced2}, or RDN~\cite{zhang2018residual2}, both of which drop their upsampling layers, and their output channel dimensions $d_e=64$. We employ the multi-head attention scheme in \cite{vaswani2017attention2} by dividing the queries, keys and values into $n_{heads}$ parts with each of dimension $d_z/n_{heads}$. In our implementation, $d_z=256,n_{heads}=16$, yielding 16-dimensional output values. We only use two iterations $(T=2)$ in the kernel integral operator, which already outperforms previous works, while keeping the running time advantage. Note that we utilize $1\times1$ convolutions to replace all the linear layers in SRNO, since they have a GPU-friendly data structure. 

%-------------------------------------------------------------------------

\section{Random vs. Sequential sampling}
For a single batch, we crop $B$ patches of sizes $\{128r^{(i)}\times128r^{(i)}\}_{i=1}^B$ from the HR training images (one per each). The LR counterparts are downsampled using bicubic interpolation with the corresponding $r^{(i)}$. In order to keep the consistent dimensions of the LR patches, sharing a common supervisory HR signal in a single batch, we sample $128^2$ HR pixels and calculate the corresponding fractional coordinates on the coarse grid associated with $r^{(i)}$. Figure \ref{fig:samplem} shows two different ways to sample function values. Experiments, in Tab.\ref{tab:11}, Tab.\ref{tab:22} and Fig.\ref{fig:vs}, verify that the random sampling method achieves better performance than the sequential sampling. These results show that using random sampling method can capture a more comprehensive representation for an image function, which is attributed to the fact that random coordinates, as demonstrated in Fig.\ref{fig:coord}, contain some extra and useful high-frequency information for SR reconstruction.  
%-------------------------------------------------------------------------

\section{Additional Results}
\begin{figure*}
    \centering
    \tabcolsep=0.02in
    \begin{tabular}{cccccc}
        \raisebox{0.1in}{\rotatebox{90}{Urban100($\times8$)}}
        \includegraphics[width=1.348in, height=1in]{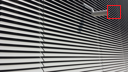}&
        \includegraphics[width=1in]{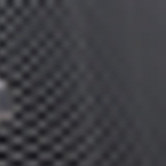} &
        \includegraphics[width=1in]{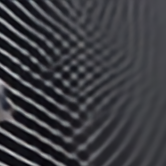} & \includegraphics[width=1in]{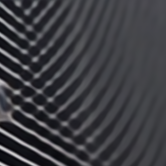} &
        \includegraphics[width=1in]{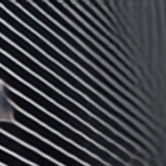} & \includegraphics[width=1in]{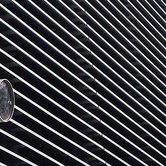} \\
        
        \raisebox{0.075in}{\rotatebox{90}{Urban100($\times4$)}}
        \includegraphics[width=1.348in, height=1in]{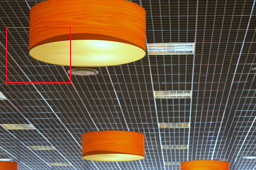}&
        \includegraphics[width=1in]{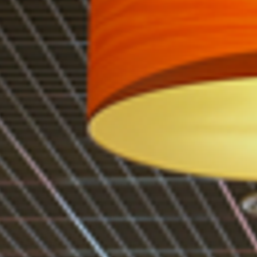} &
        \includegraphics[width=1in]{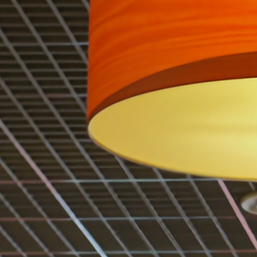} & \includegraphics[width=1in]{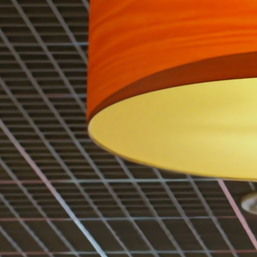} &
        \includegraphics[width=1in]{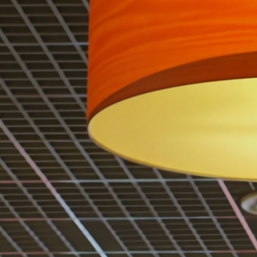} & \includegraphics[width=1in]{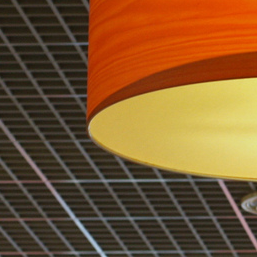} \\
        
        \raisebox{0.075in}{\rotatebox{90}{Urban100($\times8$)}}
        \includegraphics[width=1.348in, height=1in]{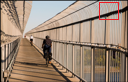}&
        \includegraphics[width=1in]{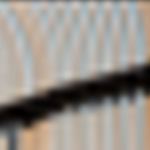} &
        \includegraphics[width=1in]{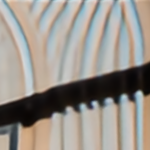} & \includegraphics[width=1in]{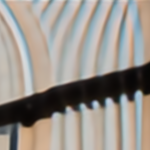} &
        \includegraphics[width=1in]{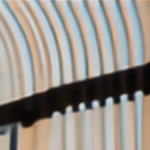} & \includegraphics[width=1in]{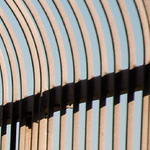} \\
        
        \raisebox{0.075in}{\rotatebox{90}{Urban100($\times4$)}}
        \includegraphics[width=1.348in, height=1in]{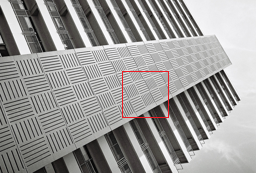}&
        \includegraphics[width=1in]{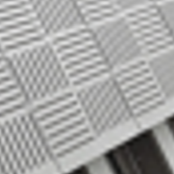} &
        \includegraphics[width=1in]{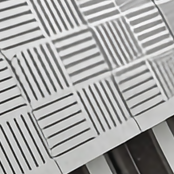} & \includegraphics[width=1in]{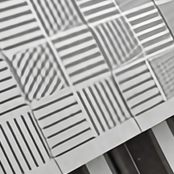} &
        \includegraphics[width=1in]{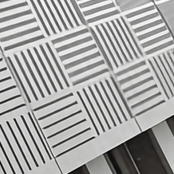} & \includegraphics[width=1in]{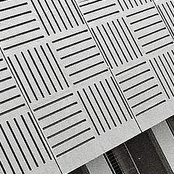} \\
        
        \raisebox{0.075in}{\rotatebox{90}{Set14($\times2.3$)}}
        \includegraphics[width=1.348in, height=1in]{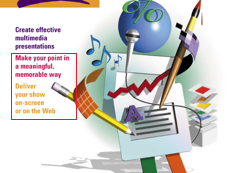}&
        \includegraphics[width=1in]{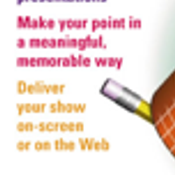} &
        \includegraphics[width=1in]{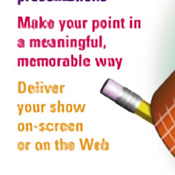} & \includegraphics[width=1in]{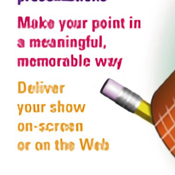} &
        \includegraphics[width=1in]{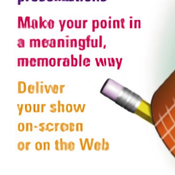} & \includegraphics[width=1in]{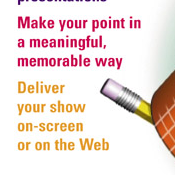} \\
        
        \raisebox{0.075in}{\rotatebox{90}{Set14($\times7.5$)}}
        \includegraphics[width=1.348in, height=1in]{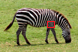}&
        \includegraphics[width=1in]{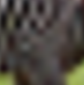} &
        \includegraphics[width=1in]{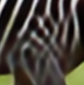} & \includegraphics[width=1in]{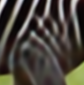} &
        \includegraphics[width=1in]{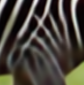} & \includegraphics[width=1in]{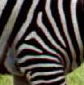} \\
        
        \begin{tabular}{c} LR Image \end{tabular} & Bicubic & LIIF~\cite{chen2021learning2} & LTE~\cite{lee2022local2} & SRNO (ours) & GT
    \end{tabular}
    \caption{\textbf{Visual comparison on other zero-shot SR}. The boxes in the first column indicate the areas that the close-ups on the right display. All methods are trained with continuous random scales in $\times1$--$\times4$. RDN is used as the encoder for all methods.}
    %\vspace{-0.5em}
    \label{fig:v}
\end{figure*}

We further compare our SRNO to LIIF, LTE on several images in Fig.\ref{fig:v}. It can be ovserved from the zoom-in regions that our SRNO consistently produces clearer and finer details than others.

%Figure \ref{fig:t} shows some super-resolution examples of scene text images from \cite{wang2020scene}.

\begin{table*}
    \centering
    \scalebox{1}{
    \begin{tabular}{c|ccc|ccccc}
        \multirow{2}{*}{Method} & \multicolumn{3}{c|}{In-distribution} & \multicolumn{5}{c}{Out-of-distribution} \\
        & $\times$2 & $\times$3 & $\times$4 & $\times$6 & $\times$12 & $\times$18 & $\times$24 & $\times$30 \\
        \hline
        Bicubic & 31.01 & 28.22 & 26.66 & 24.82 & 22.27 & 21.00 & 20.19 & 19.59 \\
        EDSR-baseline-LTE~\cite{lee2022local2} & 34.72 & 31.02 & 29.04 &
        26.81 & 23.78 & 22.23 & 21.24 & 20.53 \\
        EDSR-baseline-SRNO & \textbf{34.85} & \textbf{31.11} & \textbf{29.16} &
        \textbf{26.90} & \textbf{23.84} & \textbf{22.29} & \textbf{21.27} & \textbf{20.56} \\
        EDSR-baseline-SRNO (-s) & 34.79 & 31.07 & 29.09 & 26.84 & 23.80 & 22.26 & 21.24 & 20.54 \\
    \end{tabular}}
    %\vspace{-10pt}
    \caption{\textbf{Random vs. Sequential sampling of SRNO on DIV2K validation set (PSNR (dB))}. The best performance are bolded. All methods are trained with continuous random scales uniformly sampled in $\times1$--$\times4$. -s refers to using sequential sampling when training.}
    \label{tab:11}
\end{table*}

\begin{table*}
    \centering
    \scalebox{1}{
    \begin{tabular}{c|c|ccc|cc}
        \multirow{2}{*}{Dataset} & \multirow{2}{*}{Method} & \multicolumn{3}{c|}{In-distribution} & \multicolumn{2}{c}{Out-of-distribution} \\
        & & $\times$2 & $\times$3 & $\times$4 & $\times$6 & $\times$8 \\
        \hline
        \multirow{3}{*}{Set5} & EDSR-baseline-LTE~\cite{lee2022local2} & 38.03 & 34.48 & 32.27 & 28.96 & 27.04 \\
        & EDSR-baseline-SRNO & \textbf{38.15} & \textbf{34.53} & \textbf{32.39} & \textbf{29.06} & \textbf{27.06} \\
        & EDSR-baseline-SRNO (-s) & 38.12 & 34.50 & 32.37 & 28.96 & 27.04 \\
        \hline
        \multirow{3}{*}{Set14} & EDSR-baseline-LTE~\cite{lee2022local2} & 33.71 & 30.41 & 28.67 & 26.49 & 24.98 \\
        & EDSR-baseline-SRNO & \textbf{33.83} & \textbf{30.50} & \textbf{28.79} & \textbf{26.55} & \textbf{25.05} \\
        & EDSR-baseline-SRNO (-s) & 33.79 & 30.42 & 28.71 & 26.52 & 25.00 \\
        \hline
        \multirow{3}{*}{B100} & EDSR-baseline-LTE~\cite{lee2022local2} & 32.22 & 29.15 & 27.63 & 25.87 & 24.83 \\
        & EDSR-baseline-SRNO & \textbf{32.28} & \textbf{29.20} & \textbf{27.68} & \textbf{25.91} & \textbf{24.88} \\
         & EDSR-baseline-SRNO (-s) & 32.25 & 29.18 & 27.65 & 25.90 & 24.85 \\
        \hline
        \multirow{3}{*}{Urban100} & EDSR-baseline-LTE~\cite{lee2022local2} & 32.29 & 28.32 & 26.25 & 23.84 & 22.52 \\
        & EDSR-baseline-SRNO & \textbf{32.60} & \textbf{28.56} & \textbf{26.50} & \textbf{24.08} & \textbf{22.70} \\
        & EDSR-baseline-SRNO (-s) & 32.50 & 28.51 & 26.39 & 23.95 & 22.61 \\
    \end{tabular}}
    \caption{\textbf{Random vs. Sequential sampling of SRNO on benchmark datasets (PSNR (dB))}. The best performances are in bold. All methods are trained with continuous random scales uniformly sampled in $\times1$--$\times4$.  -s refers to using sequential sampling when training.}
    \label{tab:22}
    %\vspace{-0.5em}
\end{table*}

%-------------------------------------------------------------------------

%\section{Discussion}
%%%%%%%%% REFERENCES
\clearpage

\begin{spacing}{0.923}

\end{spacing}
\end{appendices}


{\small
\begin{thebibliography}{10}\itemsep=-1pt

    \bibitem{agustsson2017ntire}
    Eirikur Agustsson and Radu Timofte.
    \newblock Ntire 2017 challenge on single image super-resolution: Dataset and
      study.
    \newblock In {\em Proceedings of the IEEE conference on computer vision and
      pattern recognition workshops}, pages 126--135, 2017.
    
    \bibitem{bevilacqua2012low}
    Marco Bevilacqua, Aline Roumy, Christine Guillemot, and Marie line
      Alberi~Morel.
    \newblock Low-complexity single-image super-resolution based on nonnegative
      neighbor embedding.
    \newblock In {\em Proceedings of the British Machine Vision Conference}, pages
      135.1--135.10. BMVA Press, 2012.
    
    \bibitem{cao2021choose}
    Shuhao Cao.
    \newblock Choose a transformer: Fourier or galerkin.
    \newblock {\em Advances in Neural Information Processing Systems},
      34:24924--24940, 2021.
    
    \bibitem{chen2021pre}
    Hanting Chen, Yunhe Wang, Tianyu Guo, Chang Xu, Yiping Deng, Zhenhua Liu, Siwei
      Ma, Chunjing Xu, Chao Xu, and Wen Gao.
    \newblock Pre-trained image processing transformer.
    \newblock In {\em Proceedings of the IEEE/CVF Conference on Computer Vision and
      Pattern Recognition}, pages 12299--12310, 2021.
    
    \bibitem{chen2021learning}
    Yinbo Chen, Sifei Liu, and Xiaolong Wang.
    \newblock Learning continuous image representation with local implicit image
      function.
    \newblock In {\em Proceedings of the IEEE/CVF conference on computer vision and
      pattern recognition}, pages 8628--8638, 2021.
    
    \bibitem{dosovitskiy2020image}
    Alexey Dosovitskiy, Lucas Beyer, Alexander Kolesnikov, Dirk Weissenborn,
      Xiaohua Zhai, Thomas Unterthiner, Mostafa Dehghani, Matthias Minderer, Georg
      Heigold, Sylvain Gelly, et~al.
    \newblock An image is worth 16x16 words: Transformers for image recognition at
      scale.
    \newblock In {\em International Conference on Learning Representations}, 2020.
    
    \bibitem{ern2004theory}
    Alexandre Ern and Jean-Luc Guermond.
    \newblock {\em Theory and practice of finite elements}, volume 159.
    \newblock Springer, 2004.
    
    \bibitem{gu2021interpreting}
    Jinjin Gu and Chao Dong.
    \newblock Interpreting super-resolution networks with local attribution maps.
    \newblock In {\em Proceedings of the IEEE/CVF Conference on Computer Vision and
      Pattern Recognition}, pages 9199--9208, 2021.
    
    \bibitem{guibas2021efficient}
    John Guibas, Morteza Mardani, Zongyi Li, Andrew Tao, Anima Anandkumar, and
      Bryan Catanzaro.
    \newblock Efficient token mixing for transformers via adaptive fourier neural
      operators.
    \newblock In {\em International Conference on Learning Representations}, 2021.
    
    \bibitem{gupta2021multiwavelet}
    Gaurav Gupta, Xiongye Xiao, and Paul Bogdan.
    \newblock Multiwavelet-based operator learning for differential equations.
    \newblock {\em Advances in Neural Information Processing Systems},
      34:24048--24062, 2021.
    
    \bibitem{hu2019meta}
    Xuecai Hu, Haoyuan Mu, Xiangyu Zhang, Zilei Wang, Tieniu Tan, and Jian Sun.
    \newblock Meta-sr: A magnification-arbitrary network for super-resolution.
    \newblock In {\em Proceedings of the IEEE/CVF conference on computer vision and
      pattern recognition}, pages 1575--1584, 2019.
    
    \bibitem{huang2015single}
    Jia-Bin Huang, Abhishek Singh, and Narendra Ahuja.
    \newblock Single image super-resolution from transformed self-exemplars.
    \newblock In {\em Proceedings of the IEEE conference on computer vision and
      pattern recognition}, pages 5197--5206, 2015.
    
    \bibitem{hwang2022solving}
    Rakhoon Hwang, Jae~Yong Lee, Jin~Young Shin, and Hyung~Ju Hwang.
    \newblock Solving pde-constrained control problems using operator learning.
    \newblock {\em Proceedings of the AAAI Conference on Artificial Intelligence},
      36(4):4504--4512, 2022.
    
    \bibitem{islam2019much}
    Md~Amirul Islam, Sen Jia, and Neil~DB Bruce.
    \newblock How much position information do convolutional neural networks
      encode?
    \newblock In {\em International Conference on Learning Representations}, 2019.
    
    \bibitem{jiang2020meshfreeflownet}
    Chiyu"~Max" Jiang, Soheil Esmaeilzadeh, Kamyar Azizzadenesheli, Karthik
      Kashinath, Mustafa Mustafa, Hamdi~A Tchelepi, Philip Marcus, and Anima
      Anandkumar.
    \newblock Meshfreeflownet: a physics-constrained deep continuous space-time
      super-resolution framework.
    \newblock In {\em Proceedings of the International Conference for High
      Performance Computing, Networking, Storage and Analysis}, pages 1--15, 2020.
    
    \bibitem{kingma2015adam}
    Diederik~P Kingma and Jimmy Ba.
    \newblock Adam: A method for stochastic optimization.
    \newblock In {\em ICLR (Poster)}, 2015.
    
    \bibitem{kovachki2021neural}
    Nikola Kovachki, Zongyi Li, Burigede Liu, Kamyar Azizzadenesheli, Kaushik
      Bhattacharya, Andrew Stuart, and Anima Anandkumar.
    \newblock Neural operator: Learning maps between function spaces.
    \newblock {\em arXiv preprint arXiv:2108.08481}, 2021.
    
    \bibitem{lee2022local}
    Jaewon Lee and Kyong~Hwan Jin.
    \newblock Local texture estimator for implicit representation function.
    \newblock In {\em Proceedings of the IEEE/CVF Conference on Computer Vision and
      Pattern Recognition}, pages 1929--1938, 2022.
    
    \bibitem{li2020fourier}
    Zongyi Li, Nikola~Borislavov Kovachki, Kamyar Azizzadenesheli, Kaushik
      Bhattacharya, Andrew Stuart, Anima Anandkumar, et~al.
    \newblock Fourier neural operator for parametric partial differential
      equations.
    \newblock In {\em International Conference on Learning Representations}, 2020.
    
    \bibitem{9607618}
    Jingyun Liang, Jiezhang Cao, Guolei Sun, Kai Zhang, Luc Van~Gool, and Radu
      Timofte.
    \newblock Swinir: Image restoration using swin transformer.
    \newblock In {\em 2021 IEEE/CVF International Conference on Computer Vision
      Workshops (ICCVW)}, pages 1833--1844, 2021.
    
    \bibitem{lim2017enhanced}
    Bee Lim, Sanghyun Son, Heewon Kim, Seungjun Nah, and Kyoung Mu~Lee.
    \newblock Enhanced deep residual networks for single image super-resolution.
    \newblock In {\em Proceedings of the IEEE conference on computer vision and
      pattern recognition workshops}, pages 136--144, 2017.
    
    \bibitem{liu2021swin}
    Ze Liu, Yutong Lin, Yue Cao, Han Hu, Yixuan Wei, Zheng Zhang, Stephen Lin, and
      Baining Guo.
    \newblock Swin transformer: Hierarchical vision transformer using shifted
      windows.
    \newblock In {\em Proceedings of the IEEE/CVF International Conference on
      Computer Vision}, pages 10012--10022, 2021.
    
    \bibitem{lu2021learning}
    Lu Lu, Pengzhan Jin, Guofei Pang, Zhongqiang Zhang, and George~Em Karniadakis.
    \newblock Learning nonlinear operators via deeponet based on the universal
      approximation theorem of operators.
    \newblock {\em Nature Machine Intelligence}, 3(3):218--229, 2021.
    
    \bibitem{magid2021dynamic}
    Salma~Abdel Magid, Yulun Zhang, Donglai Wei, Won-Dong Jang, Zudi Lin, Yun Fu,
      and Hanspeter Pfister.
    \newblock Dynamic high-pass filtering and multi-spectral attention for image
      super-resolution.
    \newblock In {\em Proceedings of the IEEE/CVF International Conference on
      Computer Vision}, pages 4288--4297, 2021.
    
    \bibitem{martin2001database}
    David Martin, Charless Fowlkes, Doron Tal, and Jitendra Malik.
    \newblock A database of human segmented natural images and its application to
      evaluating segmentation algorithms and measuring ecological statistics.
    \newblock In {\em Proceedings Eighth IEEE International Conference on Computer
      Vision. ICCV 2001}, volume~2, pages 416--423. IEEE, 2001.
    
    \bibitem{park2019deepsdf}
    Jeong~Joon Park, Peter Florence, Julian Straub, Richard Newcombe, and Steven
      Lovegrove.
    \newblock Deepsdf: Learning continuous signed distance functions for shape
      representation.
    \newblock In {\em Proceedings of the IEEE/CVF conference on computer vision and
      pattern recognition}, pages 165--174, 2019.
    
    \bibitem{pathak2022fourcastnet}
    Jaideep Pathak, Shashank Subramanian, Peter Harrington, Sanjeev Raja, Ashesh
      Chattopadhyay, Morteza Mardani, Thorsten Kurth, David Hall, Zongyi Li, Kamyar
      Azizzadenesheli, et~al.
    \newblock Fourcastnet: A global data-driven high-resolution weather model using
      adaptive fourier neural operators.
    \newblock {\em arXiv preprint arXiv:2202.11214}, 2022.
    
    \bibitem{rudin1991functional}
    Walter Rudin.
    \newblock Functional analysis, mcgrawhill.
    \newblock {\em Inc, New York}, 45:46, 1991.
    
    \bibitem{shi2016real}
    Wenzhe Shi, Jose Caballero, Ferenc Husz{\'a}r, Johannes Totz, Andrew~P Aitken,
      Rob Bishop, Daniel Rueckert, and Zehan Wang.
    \newblock Real-time single image and video super-resolution using an efficient
      sub-pixel convolutional neural network.
    \newblock In {\em Proceedings of the IEEE conference on computer vision and
      pattern recognition}, pages 1874--1883, 2016.
    
    \bibitem{sitzmann2020implicit}
    Vincent Sitzmann, Julien Martel, Alexander Bergman, David Lindell, and Gordon
      Wetzstein.
    \newblock Implicit neural representations with periodic activation functions.
    \newblock {\em Advances in Neural Information Processing Systems},
      33:7462--7473, 2020.
    
    \bibitem{son2021srwarp}
    Sanghyun Son and Kyoung~Mu Lee.
    \newblock Srwarp: Generalized image super-resolution under arbitrary
      transformation.
    \newblock In {\em Proceedings of the IEEE/CVF conference on computer vision and
      pattern recognition}, pages 7782--7791, 2021.
    
    \bibitem{tancik2020fourier}
    Matthew Tancik, Pratul Srinivasan, Ben Mildenhall, Sara Fridovich-Keil, Nithin
      Raghavan, Utkarsh Singhal, Ravi Ramamoorthi, Jonathan Barron, and Ren Ng.
    \newblock Fourier features let networks learn high frequency functions in low
      dimensional domains.
    \newblock {\em Advances in Neural Information Processing Systems},
      33:7537--7547, 2020.
    
    \bibitem{timofte2017ntire}
    Radu Timofte, Eirikur Agustsson, Luc Van~Gool, Ming-Hsuan Yang, and Lei Zhang.
    \newblock Ntire 2017 challenge on single image super-resolution: Methods and
      results.
    \newblock In {\em Proceedings of the IEEE conference on computer vision and
      pattern recognition workshops}, pages 114--125, 2017.
    
    \bibitem{touvron2021training}
    Hugo Touvron, Matthieu Cord, Matthijs Douze, Francisco Massa, Alexandre
      Sablayrolles, and Herv{\'e} J{\'e}gou.
    \newblock Training data-efficient image transformers \& distillation through
      attention.
    \newblock In {\em International Conference on Machine Learning}, pages
      10347--10357. PMLR, 2021.
    
    \bibitem{vapnik1998statistical}
    V Vapnik.
    \newblock Statistical learning theory wiley-interscience.
    \newblock {\em New York}, 1998.
    
    \bibitem{vaswani2017attention}
    Ashish Vaswani, Noam Shazeer, Niki Parmar, Jakob Uszkoreit, Llion Jones,
      Aidan~N Gomez, {\L}ukasz Kaiser, and Illia Polosukhin.
    \newblock Attention is all you need.
    \newblock {\em Advances in neural information processing systems}, 30, 2017.
    
    \bibitem{wang2021learning}
    Longguang Wang, Yingqian Wang, Zaiping Lin, Jungang Yang, Wei An, and Yulan
      Guo.
    \newblock Learning a single network for scale-arbitrary super-resolution.
    \newblock In {\em Proceedings of the IEEE/CVF international conference on
      computer vision}, pages 4801--4810, 2021.
    
    \bibitem{yang2021implicit}
    Jingyu Yang, Sheng Shen, Huanjing Yue, and Kun Li.
    \newblock Implicit transformer network for screen content image continuous
      super-resolution.
    \newblock {\em Advances in Neural Information Processing Systems},
      34:13304--13315, 2021.
    
    \bibitem{zeyde2010single}
    Roman Zeyde, Michael Elad, and Matan Protter.
    \newblock On single image scale-up using sparse-representations.
    \newblock In {\em International conference on curves and surfaces}, pages
      711--730. Springer, 2010.
    
    \bibitem{zhang2018image}
    Yulun Zhang, Kunpeng Li, Kai Li, Lichen Wang, Bineng Zhong, and Yun Fu.
    \newblock Image super-resolution using very deep residual channel attention
      networks.
    \newblock In {\em Proceedings of the European conference on computer vision
      (ECCV)}, pages 286--301, 2018.
    
    \bibitem{zhang2018residual}
    Yulun Zhang, Yapeng Tian, Yu Kong, Bineng Zhong, and Yun Fu.
    \newblock Residual dense network for image super-resolution.
    \newblock In {\em Proceedings of the IEEE conference on computer vision and
      pattern recognition}, pages 2472--2481, 2018.
    
    \end{thebibliography}
    
}


\small\begin{thebibliography}{1}\itemsep=-1pt

\bibitem{cao2021choose2}
Shuhao Cao.
\newblock Choose a transformer: Fourier or galerkin.
\newblock {\em Advances in Neural Information Processing Systems},
  34:24924--24940, 2021.

\bibitem{chen2021learning2}
Yinbo Chen, Sifei Liu, and Xiaolong Wang.
\newblock Learning continuous image representation with local implicit image
  function.
\newblock In {\em Proceedings of the IEEE/CVF conference on computer vision and
  pattern recognition}, pages 8628--8638, 2021.

\bibitem{ciarlet2013linear2}
Philippe~G Ciarlet.
\newblock {\em Linear and nonlinear functional analysis with applications},
  volume 130.
\newblock Siam, 2013.

\bibitem{ern2004theory2}
Alexandre Ern and Jean-Luc Guermond.
\newblock {\em Theory and practice of finite elements}, volume 159.
\newblock Springer, 2004.

\bibitem{lee2022local2}
Jaewon Lee and Kyong~Hwan Jin.
\newblock Local texture estimator for implicit representation function.
\newblock In {\em Proceedings of the IEEE/CVF Conference on Computer Vision and
  Pattern Recognition}, pages 1929--1938, 2022.

\bibitem{lim2017enhanced2}
Bee Lim, Sanghyun Son, Heewon Kim, Seungjun Nah, and Kyoung Mu~Lee.
\newblock Enhanced deep residual networks for single image super-resolution.
\newblock In {\em Proceedings of the IEEE conference on computer vision and
  pattern recognition workshops}, pages 136--144, 2017.

\bibitem{vaswani2017attention2}
Ashish Vaswani, Noam Shazeer, Niki Parmar, Jakob Uszkoreit, Llion Jones,
  Aidan~N Gomez, {\L}ukasz Kaiser, and Illia Polosukhin.
\newblock Attention is all you need.
\newblock {\em Advances in neural information processing systems}, 30, 2017.

\bibitem{zhang2018residual2}
Yulun Zhang, Yapeng Tian, Yu Kong, Bineng Zhong, and Yun Fu.
\newblock Residual dense network for image super-resolution.
\newblock In {\em Proceedings of the IEEE conference on computer vision and
  pattern recognition}, pages 2472--2481, 2018.

\end{thebibliography}
\end{document}